\documentclass[11pt]{article}

\usepackage[final]{acl}

\usepackage{times}
\usepackage{latexsym}

\usepackage[T1]{fontenc}
\usepackage[T5]{fontenc}

\usepackage[utf8]{inputenc}

\usepackage{microtype}

\usepackage{inconsolata}

\usepackage{graphicx}
\usepackage{booktabs}
\usepackage{rotating}
\usepackage[most]{tcolorbox}
\usepackage{subcaption}
\usepackage{lipsum}
\usepackage{listings}
\usepackage{xcolor}
\usepackage{makecell}
\usepackage{pifont}
\usepackage{array}
\usepackage{multirow, multicol}
\tcbuselibrary{listings,breakable}

\newcommand{\cmark}{\textcolor{green!60!black}{\ding{51}}}
\newcommand{\xmark}{\textcolor{red!70!black}{\ding{55}}}

\usepackage{tcolorbox}
\tcbuselibrary{listings, breakable, skins}

\newtcblisting{promptbox}[2][]{
    colback=gray!5!white,       
    colframe=gray!75!black,     
    fonttitle=\bfseries,        
    title=#2,                   
    breakable,                  
    boxrule=0.5pt,              
    left=2mm, right=2mm, top=2mm, bottom=2mm, 
    listing only,               
    listing options={           
        basicstyle=\ttfamily\scriptsize, 
        breaklines=true,                 
        breakatwhitespace=true,
        keepspaces=true,                 
        showstringspaces=false
    },
    #1 
}

\title{C$^3$PO: Evaluating Cross-Modal Composition and Counterfactual Performance in Omnimodal Models}

\author{
 \textbf{Swapnanil Mukherjee\textsuperscript{1}},
 \textbf{Agyeya Negi\textsuperscript{2}},
 \textbf{Tanuja Ganu\textsuperscript{1}},
 \textbf{Ponnurangam Kumaraguru\textsuperscript{2}}
\\
\\
 \textsuperscript{1}Microsoft Research India,
 \textsuperscript{2}IIIT Hyderabad, India
\\
 \small{
   \textbf{Correspondence:} \href{mailto:email@domain}{swapnanil.mukherjee12@gmail.com}
 }
}

\begin{document}
\maketitle
\begin{abstract}
Current Multimodal Large Language Models (MLLMs) can process diverse sensory inputs, yet their reasoning remains heavily biased toward a dominant modality, resulting in brittle cross-modal reasoning. We introduce C$^3$PO, a benchmark of 3,404 samples spanning video, audio, image, and text, evaluating two abilities: information composition (fusing dispersed evidence) and counterfactual conflict (resolving deliberate contradictions). C$^3$PO's paired IC/CC structure and four-tier design enable targeted diagnosis of when and why cross-modal reasoning fails. Built through a fully automatic pipeline using 25 logically grounded templates, C³PO reveals that while humans achieve 88.64\% accuracy, the best model (Gemini-3.1-Pro) reaches only 73.17\%, with open-source models collapsing under conflict. Through attention probes, we find 86–95\% of failures stem from modality dominance: models commit to one modality while ignoring contradictory evidence, concentrating 87–95\% of attention on text. Mid-layer attention entropy predicts correctness—sustained exploration succeeds, premature collapse fails. The 56-point accuracy gap between equally complex templates reveals that performance depends on modalities' structural roles in conflict resolution, not combinations. These findings show multimodal perception does not guarantee robust reasoning; architectures must enable sustained cross-modal attention to avoid premature commitment.

\end{abstract}

\section{Introduction}
    \label{sec:intro}
    AI models have experienced a rapid evolution, transitioning rapidly from unimodal architectures to powerful Multimodal Large Language Models (MLLMs) capable of processing textual, visual, and auditory data simultaneously. As these architectures advanced to integrate diverse sensory streams, the paradigms for evaluating them evolved concurrently. Early assessments relied on simple tasks like captioning or question-answering, but the field has progressively shifted towards complex and multi-disciplinary audio-visual reasoning tasks \cite{li2025omnivideo, jiang2025fysicsworld}. However, despite the impressive improvement in the multimodal capabilities of such models, they are still plagued by ``modality bias''. Modality bias occurs when a model disproportionately relies on or `prefers' a single dominant modality (most frequently the pre-trained language prior), while underutilizing, ignoring, or misinterpreting accompanying inputs from other modalities \cite{zheng2025mllms, wang2025audio, cai2025diagnosing}.

    \begin{figure}
        \centering
        \includegraphics[width=\linewidth]{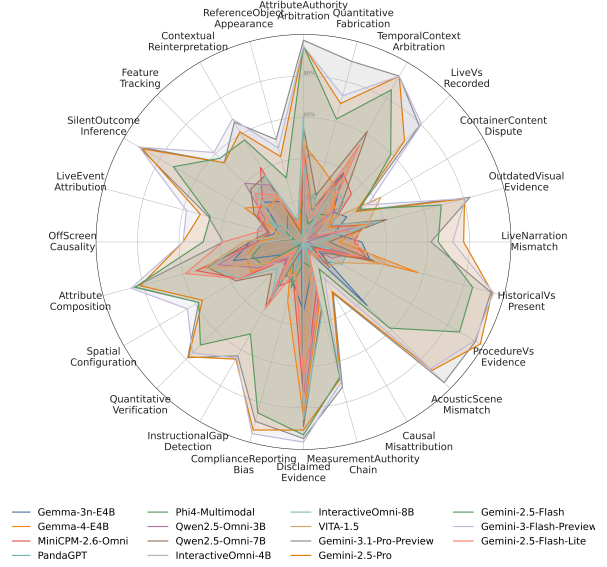}
        \caption{Template-wise breakdown of performance for leading MLLMs on the \textbf{C$^3$PO} dataset.}
        \label{fig:radar_acc_graph}
    \end{figure}

    This problem stems from the architectural and training choices underlying these models. The bias largely arises from dataset imbalances, where language data dominates pre-training while visual and auditory data remain comparatively limited, producing skewed learning dynamics \cite{leng2025curse, zheng2025mllms, cai2025diagnosing, park2025assessing}. Architectural asymmetry further exacerbates this issue: modern MLLMs use large pre-trained language encoders as the backbone, while image and audio encoders are incorporated through fine-tuning or lightweight modality-specific adapters \cite{phi4, vita1.5}. Because these modules are substantially smaller than the language model, the text modality often dominates at the expense of less mature visual and acoustic components \cite{gemmateam2025gemma3technicalreport}. Current training objectives also fail to enforce balanced cross-modal alignment, encouraging reliance on linguistic shortcuts and priors \cite{zheng2025mllms, yang-etal-2025-audio-centric}. Prior work further shows that multimodal training itself often converges toward dependence on a dominant modality, even under more balanced settings \cite{wang2020makes, pmlr-v162-huang22e}.

    The lack of cross-modal competence becomes evident when modalities present conflicting, or asynchronous information, requiring reasoning about information spread across multiple modalities \cite{leng2025curse}. Existing evaluation protocols largely rely on manually curated datasets that are expensive to scale, limited in data and task diversity, and biased toward simple perception or multiple-choice questions solvable through unimodal priors. Although recent benchmarks attempt to address these limitations, they retain key shortcomings. The majority of current datasets (such as  WorldSense\cite{hong2025worldsense}, MAVERIX \cite{xie2026maverix}, OmniVideoBench \cite{li2025omnivideo} etc.) restrict themselves to tri-modal (e.g., Video+Audio+Text) or fewer inputs. FysicsWorld \cite{jiang2025fysicsworld} includes all four modalities but requires intensive human-LLM collaborative review, limiting scalability. Moreover, benchmarks like MMA-Bench \cite{chen2025modalities} and DAVE \cite{radevski2025dave} evaluate simple modality contradictions, but lack the task diversity needed to assess both counterfactual reasoning and information composition across all four input modalities in real-world settings. Thus, our contributions in this paper are:
    \begin{enumerate}  
        \item We introduce a new benchmark dataset, \textbf{C$^3$PO}, composed of co-located video, audio, image, and text modalities to assess the capabilities of MLLMs along two broad cognitive axes, including a novel Counterfactual Conflict category.
        \item We propose a fully automatic and scalable generation framework consisting of 25 granular task templates corresponding to real-world abilities, addressing the scalability limits of manual annotation. 
        \item We conduct a fine-grained analysis of the various failure modes of  leading MLLMs across diverse tasks involving combinations of different modalities, demonstrating the effectiveness of our benchmark in exposing the modality biases that affect even the strongest omnimodal models. 
    \end{enumerate}
    
    \begin{figure}[t]
        \centering
        \includegraphics[width=\columnwidth]{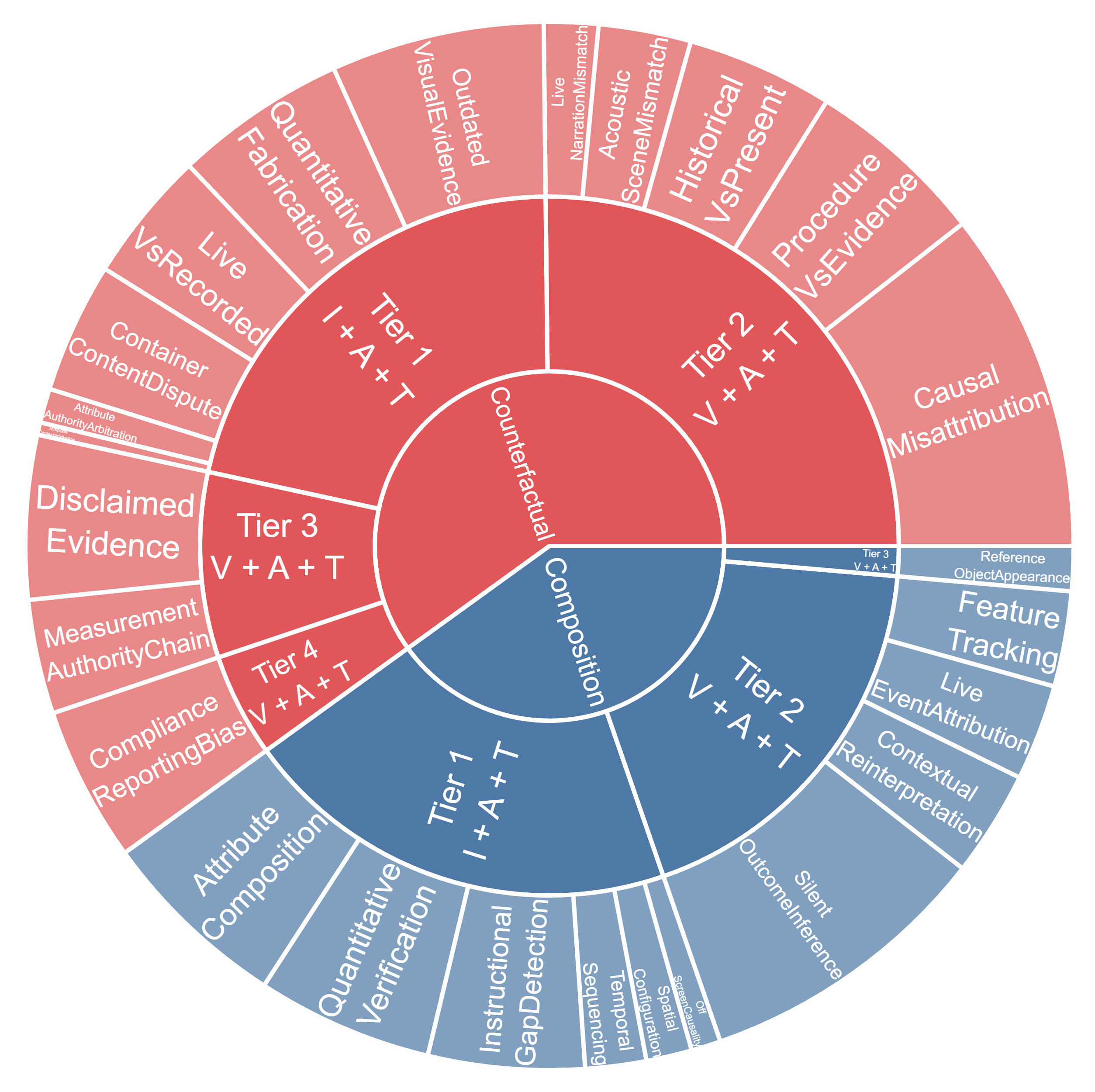}
        \caption{Hierarchical taxonomy of our C$^3$PO benchmark dataset. The two primary task categories \textbf{Information Composition} and \textbf{Counterfactual Reasoning} are subdivided into Tiers (1-4) by modality combinations. Each Tier is subdivided into 25 task templates.}
        \label{fig:sunburst}
    \end{figure}

\section{Related Work}
\label{sec:relatedwork}

    \textbf{Modality Bias and Interference in MLLMs}. Existing work has increasingly identified modality bias as a severe vulnerability in multimodal learning. \citet{leng2025curse} introduced the Curse of Multi-Modalities (CMM) benchmark, showing that MLLMs suffer hallucinations driven by overreliance on unimodal priors and spurious inter-modality correlations learned during pretraining. \citet{cai2025diagnosing} used causal perturbation-based diagnostics to expose modality interference, demonstrating that non-essential modalities heavily distort model decisions. \citet{hua2025vlm} studied Vision-Language models under conflicting inputs finding that models often exhibit ``blind faith'' in text. In the audio domain, \citet{wang2025audio} proposed MCR-BENCH for Audio-Language models and showed a widespread text bias, where models frequently disregard clear auditory evidence under contradiction. 

    \begin{figure*}[t]
        \centering
        \includegraphics[width=\textwidth]{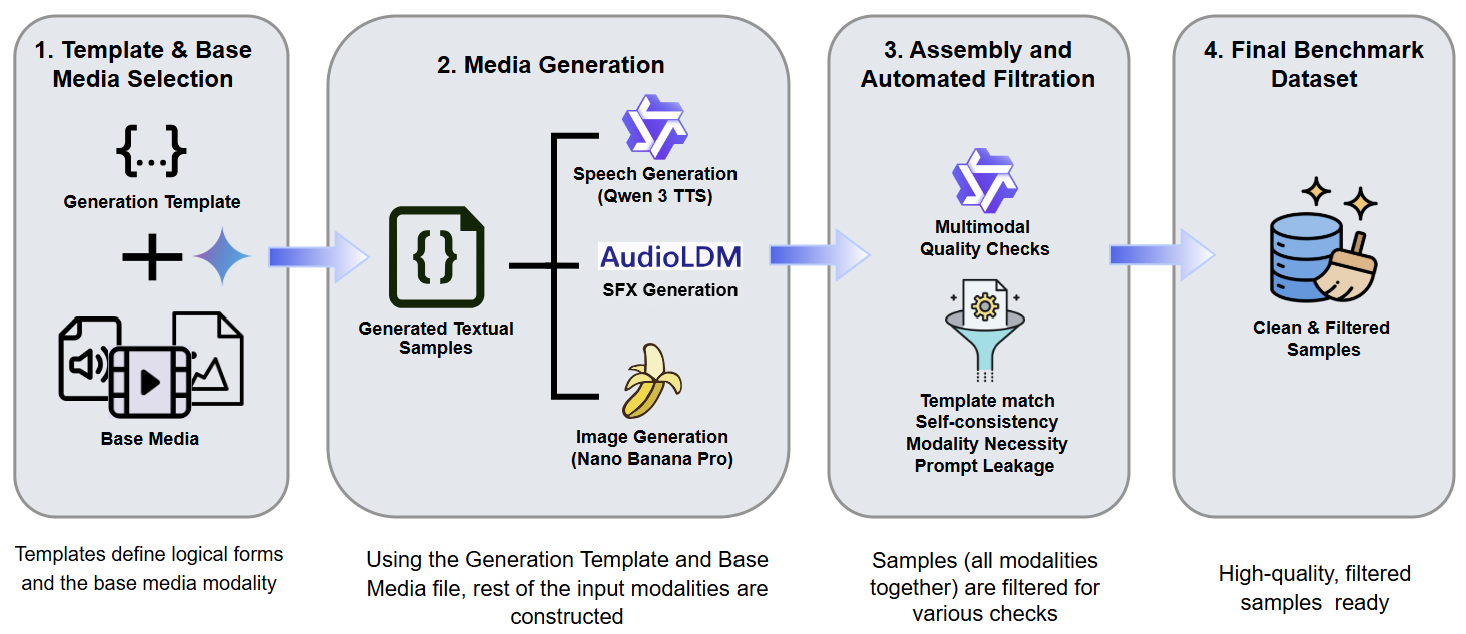}
        \caption{Our pipeline for automatic and scalable benchmark construction.}
        \label{fig:pipeline}
    \end{figure*}

    \noindent \textbf{Omnimodal Benchmarks}. Early benchmarks focused on static visual understanding and academic reasoning, including MME \cite{fu2025mme} and MMMU \cite{yue2023mmmu}. MMT-Bench \cite{pmlr-v235-ying24a} expanded this with 31,325 curated multiple-choice questions across 162 subtasks spanning multitask understanding, spatial localization, and expert knowledge. Video-MME \cite{fu2025videomme} evaluated temporal and long-form video understanding using subtitles and audio tracks across 13 task categories including perception, causal reasoning, and sentiment analysis. WorldSense \cite{hong2025worldsense} focused on omni-modal video understanding through integrated audio-visual inputs in real-world scenarios. OmniBench \cite{li2025omnibench} evaluated concurrent reasoning across visual, acoustic, and textual inputs, showing that open-source models struggle in tri-modal settings. More recent benchmarks study adversarial or conflicting conditions. DAVE \cite{radevski2025dave} and AURA \cite{galougah2025aura} examined audio-visual synchronization and cross-modal reasoning, showing that models infer spurious correlations without evidence and struggle with causal reasoning. MAVERIX \cite{xie2026maverix} introduced 8-way multiple-choice and open-ended questions across six evaluation dimensions, revealing difficulties with long videos and subtle asynchronous audio cues. JointAVBench \cite{chao2026jointav} studied multi-scene joint reasoning with three audio types, requiring correlation of auditory and visual cues across temporal and spatial contexts. \citet{wang2026xmodbench} introduced XModBench to evaluate cross-modal consistency and demonstrated strong directional imbalance and modality disparity in leading MLLMs. Finally, and SONIC-O1 \cite{radwan2026sonic} introduced bidirectional demographic-aware real-world evaluation.

\section{C$^3$PO: A Benchmark Dataset}
    \label{dataset}
    C$^3$PO consists of 3404 samples, 2137 unique video clips, 1646 unique images, 3426 unique audio files. Table \ref{tab:tier_tables} shows the distribution of samples across tiers and categories. 
    

    \subsection{Construction Methodology}
    
    \subsubsection{Data Sources}
    \label{sec:data_sources}
    To avoid the severe hallucinations and lack of physical nuance inherent in purely synthetic data, we adopt a hybrid approach grounded in authentic environments. Our automated pipeline selects real-world ``base media'' (video/image/audio) from public datasets and uses it as an anchor to synthesize the missing complementary modalities. This allows fine-grained control over the overall sample's mechanics through constrained generation while ensuring realistic and faithful task evaluation. A full list of the data sources can be found the Appendix \ref{sec:AppendixDataSources}.

    \begin{table}[t]
    \centering
    \footnotesize
    \renewcommand{\arraystretch}{1.15}
    
    \begin{tabular*}{0.92\linewidth}{@{\extracolsep{\fill}}lccc}
    \toprule
    & \makecell{\textbf{Information} \\ \textbf{Composition}} 
    & \makecell{\textbf{Counterfactual} \\ \textbf{Conflict}} 
    & \textbf{Total} \\
    \midrule
    \textbf{Tier 1} & 692 & 729  & 1133 \\
    \textbf{Tier 2} & 624 & 795  & 2295 \\
    \textbf{Tier 3} & 47  & 292  & 467 \\
    \textbf{Tier 4} & 0   & 163  & 218 \\
    \midrule
    Total & 1363 & 1979 & 3342 \\
    \bottomrule
    \end{tabular*}
    \vspace{0.5em}
    
    {\small \textbf{(a)} Distribution of samples}
    \vspace{1.5em}
    \begin{tabular*}{0.92\linewidth}{@{\extracolsep{\fill}}lcccc}
    \toprule
    & Video & Audio & Image & Text \\
    \midrule
    \textbf{Tier 1} & \xmark & \cmark$^{*}$ & \cmark$^{*}$ & \cmark \\
    \textbf{Tier 2} & \cmark$^{*}$ & \cmark & \xmark & \cmark \\
    \textbf{Tier 3} & \cmark$^{*}$ & \cmark & \cmark & \cmark \\
    \textbf{Tier 4} & \cmark$^{*}$ & \cmark & \cmark & \cmark \\
    \bottomrule
    \end{tabular*}
    
    {\small \textbf{(b)} Availability of modalities across tiers}
    \vspace{0.5em}
    \caption{Dataset distribution and modality composition across categories and tiers in C$^3$PO.$^{*}$ indicates the base modality.}
    \label{tab:tier_tables}
    \end{table}

    \begin{table*}[t]
    \small
    \centering

    \resizebox{\textwidth}{!}{
    \begin{tabular}{lccccccc}
    \toprule
    \textbf{Benchmark} & \textbf{Modalities} & \textbf{IC} & \textbf{CC} & \textbf{Tasks} & \makecell{\textbf{Source} \\ \textbf{Datasets}} & \textbf{Samples} & \makecell{\textbf{Generation} \\ \textbf{Pipeline}} \\ 
    \midrule
    Video-MME & V $+$ A $+$ T & \cmark & \xmark & 12 & 1 & 2,700 & Manual \\
        OmniBench & I $+$ A $+$ T & \cmark & \xmark & 8 & 3 & 1,142 & Manual \\
    CMM & V $+$ A $+$ T & \cmark & \cmark & 6 & 2 & $\sim$1,200 & Manual \\
    WorldSense & V $+$ A $+$ T & \cmark & \xmark & 8 & 1 & 3,172 & Manual \\
    MAVERIX & V $+$ A $+$ T & \cmark & \xmark & 7 & 5 & 2,556 & Manual \\
    OmniVideoBench & V $+$ A $+$ T & \cmark & \xmark & 13 & 2 & 1,000 & Manual \\
    Daily-Omni & V $+$ A $+$ T & \cmark & \xmark & 6 & 3 & 1,197 & Semi-Auto \\
    JointAVBench & V $+$ A $+$ T & \cmark & \xmark & 15 & 1 & 2,853 & Semi-Auto \\
    DAVE & V $+$ A $+$ T & \cmark & \cmark & 3 & 2 & 2,426 & Semi-Auto \\
    MMA-Bench & V $+$ A $+$ T & \xmark & \cmark & - & 1 & 1,316 & Semi-Auto \\
    AURA & V $+$ A $+$ T & \cmark & \xmark & 6 & 2 & 1,600+ & Auto \\
    FysicsWorld & V $+$ A $+$ I $+$ T & \cmark & \xmark & 16 & 40+ & 3,268 & Semi-Auto \\
    XModBench & V, I $+$ A $+$ T & \cmark & \xmark & 17 & 10 & 61,320 & Semi-Auto \\
    \midrule
    \textbf{C$^3$PO (Ours)} & \textbf{V $+$ A $+$ I $+$ T} & \cmark & \cmark & \textbf{25} & \textbf{8} & \textbf{3342} & \textbf{Auto} \\
    \bottomrule
    \end{tabular}
    }
    \caption{Comparison of C$^3$PO with existing multimodal and omni-modal benchmarks. \textbf{Modality}: \textbf{V}ideo, \textbf{A}udio, \textbf{I}mage, \textbf{T}ext. \textbf{IC}: Information Compostion, tests integration of dispersed information across modalities. \textbf{CC}: Counterfactual Conflict, tests counterfactual reasoning, modality interference, or deliberate misalignment.}
    \label{tab:benchmark_comparison}
    \end{table*}
     
    \subsubsection{Generation Templates}
    \label{sec:generation_templates}
    Generation templates define the relationship between the various modalities, specifying the \textit{logical form} of the task and the mechanics of solving it. They define a skeleton for instantiating the content from the base media using the logical form. The benchmark is comprised of 25 such templates\footnote{The list of all templates and their full descriptions is available in the Appendix}. An example template is provided in Appendix \ref{sec:AppendixGenerationTemplate}.
    
    \lstset{
        basicstyle=\ttfamily\footnotesize,
        breaklines=true,
        breakatwhitespace=true,
        keepspaces=true,
        showstringspaces=false
    }

     In this benchmark, we target two primary cognitive tasks: 
     \begin{enumerate} 
         \item \textbf{Information Composition}: These templates disperse the required information across various modalities, making cross-modal fusion mandatory to solve the sample correctly. For instance, in \texttt{InstructionalGapDetection}, a muted video shows a full action sequence, while the generated audio intentionally narrates only the latter half using transitional markers. The model must connect the audio's starting point to the video sequence to identify the unmentioned initial steps. 
         \item \textbf{Counterfactual Conflict}: These templates introduce cross-modal contradictions, and the model has to ignore and resolve the conflicting information. For example in \verb|CausalMisattribution|, a video shows an event triggered by a clearly visible mechanism, while the audio confidently (mis)attributes it to some other cause not present in the video. To answer correctly, the deceptive auditory claim must be ignored to trust the verifiable visual evidence. 
     \end{enumerate} 
     These templates are categorized into four tiers of increasing cognitive complexity based on their specific modality combinations. Table \ref{tab:tier_tables} details the sample distribution and modality composition across these tiers and categories\footnote{A detailed breakdown of samples by templates and tiers is available in the Appendix.}.

\subsubsection{Generation Pipeline}
To construct the samples, we employ a multi-stage pipeline consisting of candidate selection, multimodal generation, and automated filtering.\footnote{\texttt{gemini-3-pro-preview} through the Gemini API. All prompts, models, and implementation details are provided in the Appendix.}
     \paragraph{Base Media Selection} For each generation template, we select suitable source datasets depending on the base modality. Candidate media files are chosen using dataset annotations and template-specific constraints, followed by lightweight preprocessing.

    \paragraph{Multimodal Generation} Given the candidate base media, Gemini-3-Pro generates the query along with detailed descriptions of the complementary modalities to complete the sample as per the template’s logic. The corresponding image and audio files are then synthesized using state-of-the-art generative models.

    \paragraph{Self-Audit and Revision} Each generated sample undergoes an internal self-audit step during generation. Samples that can be solved through unimodal shortcuts, commonsense priors, query leakage, or ambiguous reasoning must be revised before the model outputs the sample.

\subsubsection{Filtering Process}
\label{sec:filtering_process}

    As samples are generated using ‘candidate’ base media files which may not necessarily be an ideal fit for the template’s structure, the full sample generated using that base media may be of poor quality. We apply an automated model-based filtering pipeline to ensure consistency, coherence, and multimodal dependence in the samples. 
    
     \textbf{Template Alignment} Verifies that all modalities and the query jointly satisfy the logical structure of the template. For Information Composition samples, the answer must require combining information across modalities; for Counterfactual Conflict samples, the answer must remain uniquely identifiable within a plausible conflict.
    
     \textbf{Media Verification} Ensures that media (base and generated) accurately match their textual descriptions, to mitigate any hallucinations by the generation pipeline. 
    
     \textbf{Query Leakage Detection} Checks whether the query leaks information about any modality or exposes template logic that simplifies solving.
    
     \textbf{Naturalness and Realism}: Evaluates whether the complete multimodal sample is coherent, realistic, and plausible in real-world settings.

    \subsection{Evaluation Protocol}
    The ground truth answer for each sample is a free-form string. We opted out of the multiple-choice format as it is prone to guess inflation, elimination-answering, and positional biases, ultimately overstating model performance \cite{li2025omnivideo, molfese-etal-2025-right}. Models are given all the input modalities along with the question and a standardized evaluation prompt\footnote{Prompt available in the Apppendix.}. The responses are expected to be as specific as possible (to avoid any ambiguity during evaluation) and within a few words. To accurately and scalably grade these diverse generative outputs, first, a normalized string exact match is performed between the ground truths and predictions after simple preprocessing. If this match fails, we use Qwen3-Omni as an LLM-judge to check whether the ground truth answer and predicted answer are semantically equivalent. Semantic equivalence means whether the judge can uniquely pick out the same entity/concept as the ground truth from the predicted answer in the context of the entire sample. The judge returns ``Yes'' or ``No'' and accuracy is $\frac{\#\text{Yes}}{\# \text{Total Samples}}$.

    \subsection{Results and Discussion}
    
    \begin{table*}[!t]
        \centering
        
        \small
\setlength{\tabcolsep}{7pt}
\renewcommand{\arraystretch}{1.2}

\caption{
Performance (accuracy) of various SOTA open-source and proprietary
models on the C$^3$PO dataset.
}
\label{tab:benchmark_results_main}

\begin{tabular}{l|ccc|cccc|c}
\toprule

\multirow{2}{*}{Model}
& \multicolumn{3}{c|}{Information Composition}
& \multicolumn{4}{c|}{Counterfactual Conflict}
& \multirow{2}{*}{Overall (\%)} \\

\cmidrule(lr){2-4}
\cmidrule(lr){5-8}

& Tier 1 & Tier 2 & Tier 3
& Tier 1 & Tier 2 & Tier 3 & Tier 4
& \\

\midrule

Qwen2.5-Omni-3B
& 27.87 & 26.27 & 0
& 18.11 & 8.63 & 58.22 & 2.45
& 21.53 \\

InstructOmni-4B
& 20.54 & 14.26 & 10.64
& 17.42 & 3.97 & 56.16 & 0.00
& 16.41 \\

Gemma-3n-E4B
& 18.47 & 17.95 & 6.38
& 20.03 & 16.80 & 26.03 & 19.63
& 18.83 \\

Gemma-4-E4B
& 17.68 & 26.28 & 6.38
& 28.81 & 14.47 & 61.99 & 29.45
& 25.18 \\

MiniCPM-Omni-7B
& 30.89 & 27.40 & 4.26
& 22.50 & 11.67 & 54.11 & 19.02
& 24.55 \\

VITA-1.5
& 26.11 & 20.83 & 0
& 27.71 & 17.85 & 44.18 & 15.95
& 24.07 \\

PandaGPT-7B
& 2.55 & 12.66 & 2.13
& 12.07 & 12.49 & 6.85 & 19.63
& 10.27 \\

Qwen2.5-Omni-7B
& 35.35 & 29.01 & 4.26
& 30.32 & 12.02 & 66.10 & 4.29
& 27.81 \\

InstructOmni-8B
& 32.32 & 20.99 & 10.64
& 24.83 & 5.02 & 64.73 & 1.23
& 22.57 \\

Qwen-3-Omni-Instruct
& 62.62 & 53.23 & 18.93
& 41.98 & 36.74 & 77.76 & 15.64
& 43.98 \\

\midrule

gemini-3.1-pro-preview
& 73.09 & 74.04 & \textbf{51.06}
& \textbf{74.90} & 64.99 & \textbf{85.62} & 89.57
& \textbf{73.17} \\

gemini-3-flash-preview
& \textbf{74.36} & \textbf{77.08} & 46.81
& 71.19 & 61.26 & 85.62 & \textbf{95.71}
& 72.46 \\

gemini-2.5-pro
& 73.25 & 72.76 & 42.55
& 66.67 & \textbf{65.34} & 81.15 & 93.87
& 70.99 \\

gemini-2.5-flash
& 67.68 & 62.98 & 31.91
& 59.67 & 51.69 & 82.53 & 85.28
& 62.60 \\

gemini-2.5-flash-lite
& 35.03 & 25.16 & 8.51
& 16.74 & 11.32 & 50.00 & 14.72
& 23.05 \\

\midrule

Human
& -- & -- & --
& -- & -- & -- & --
& \underline{88.64} \\

\bottomrule
\end{tabular}

\end{table*}
    
We evaluate a diverse cohort of open-source and proprietary omnimodal models on
\textbf{C$^3$PO} (Table~\ref{tab:benchmark_results_main}). For open-source models we
report accuracy averaged over three independent inference runs; API limits
restrict closed-source models to a single run. Gemini-3.1-Pro leads
at $73.17\%$, while open models
fare substantially worse, with Qwen3-Omni, the largest, scoring $43.98\%$. The performance of smaller open source models is close to that of random-guessing in a four-way MCQ format. All models perform much worse than the human baseline.

Performance is highly template-dependent. Gemini-3.1-Pro maintains broad competence across templates except CausalMisattribution and ContainerContentDispute (Fig. \ref{fig:radar_acc_graph}), while open-source models collapse on templates requiring causal or temporal reasoning. The performance spread within categories is dramatic: the hardest template (CausalMisattribution, 33\% cohort-mean accuracy) and the easiest (DisclaimedEvidence, 92\%) are both Counterfactual Conflict tasks with contradictory audio, yet differ by 56\%. This reveals that task category alone does not determine difficulty. We  analyze the relationship between template structure, task category, and what they reveal about how models resolve cross-modal conflict in Section \ref{sec:mech-template}.

\subsubsection{Composition vs.\ Conflict, and the Tier Effect}
Separating performance along C$^3$PO's two cognitive axes reveals two
patterns that prior multimodal benchmarks do expose at this granularity. Firstly, all evaluated models perform better on Information Composition (IC) than on Counterfactual Conflict (CC) when the conflict is at low tier, but the gap closes and frequently inverts at higher tiers (Table~\ref{tab:benchmark_results_main}). On IC, accuracy degrades monotonically with tier: each added modality contributes a piece of the answer that the model must combine, and the fusion burden grows. On CC, accuracy is non-monotonic in nearly all models: Tier-3 accuracy
exceeds Tier-2 accuracy by a substantial margin before collapsing at Tier 4 again. We trace this non-monotonicity to template structure in Section~\ref{sec:mech-template}.
Secondly, when models fail, they tend to fail by trusting one modality and
ignoring the other rather than by partially synthesising both. We quantify
this cohort-wide in Section~\ref{sec:mech-budget}.



\section{A Mechanistic Analysis of Performance}
\label{sec:mech}

C$^3$PO's paired Information Composition (IC) and Counterfactual Conflict (CC) structure enables targeted probing of how models respond to conflict versus complementary information. Inspired by \citet{selvakumar2026audio} We apply three lightweight interpretability probes (E1, E2, E3) to analyse: (1) whether models internally detect cross-modal conflict, and (2) how attention redistributes as modalities accumulate. E1 is an attention probe measuring attention allocation across modalities during generation. E2 is a logit-lens probe testing whether modality information is internally recoverable before decoding; and E3, a caption-fidelity probe evaluating how faithfully models can reconstruct modality content. Together, E1 and E2 distinguish \emph{perception} failures (modality information never enters the latent stream) from \emph{rendering} failures (information reaches the latent stream but not the output). Full probe definitions, protocols, and cohort details are provided in Appendix~\ref{app:mech}.


\subsection{Cross-Modal Conflict Detection: Attention as a Confusion Signal}
\label{sec:mech-conflict}

We hypothesize that if models internally detect cross-modal conflict, this should manifest as higher entropy in their attention distribution: greater uncertainty when modalities contradict than when they complement. To test this, we compute the Shannon entropy ($H_\ell = -\sum_{i \in \{\text{text}, \text{audio}, \text{image}\}}p_i \log_2 p_i$) of the attention distribution across the tokens for text, audio, and image for each sample, averaging this across the middle layers to compute a scalar \emph{mid-layer entropy} (depth ranges are provided in the Appendix). Here, we exclude videos due to computational memory constraints.
\begin{figure}[h]
\centering
\includegraphics[width=0.95\columnwidth]{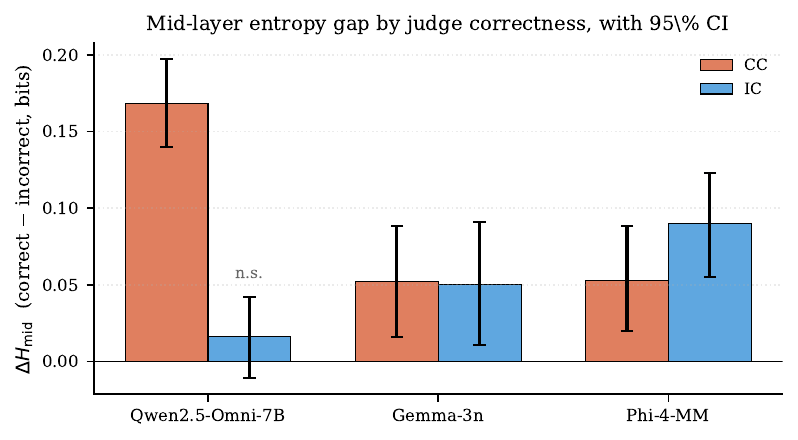}
\caption{Mid-layer between-modality Shannon entropy: difference between judge-correct and judge-incorrect samples. Five of six model-category pairs show significant positive differences (bootstrap $95\%$ CIs); higher entropy correlates with correct answers.}
\label{fig:entropy-judge}
\end{figure}

Samples marked correct by the LLM judge consistently hold higher mid-layer entropy than incorrect samples (Figure~\ref{fig:entropy-judge}). Across three models and the two task categories, five of six pairs show a statistically significant positive (the bootstrap $95\%$ confidence intervals for these five pairs strictly exclude zero) positive entropy gap between correct and incorrect predictions ($\Delta H = \overline{H}_{\text{mid}}^{\,\text{correct}} -\overline{H}_{\text{mid}}^{\,\text{incorrect}}$). When pooled using within-model $z$-normalization, the judge-correctness gap remains robustly positive under both CC ($+0.43$ standard deviations) and IC ($+0.21$ standard deviations) tasks.. 

While the \emph{direction} of the effect is consistent across the cohort, the \emph{relative strength} of the effect under CC versus IC varies by model (see Appendix for detailed bit gaps). The universal takeaway across the cohort is that premature attention concentration in the middle layers is a precursor to failure, regardless of whether the task is conflict-driven or composition-driven. The further claim that conflict resolution specifically requires sustained mid-layer breadth is strongly supported by Qwen2.5-Omni-7B, but more probed models are needed to confirm it as a cohort universal.

\subsection{Attention Budget Under Modality Pressure}
\label{sec:mech-budget}

C$^3$PO's tier structure allows us to measure how models redistribute attention as modalities accumulate. Four models support full routing analysis via E1; five models contribute to observations from E3.

\begin{figure}[h]
\centering
\includegraphics[width=\columnwidth]{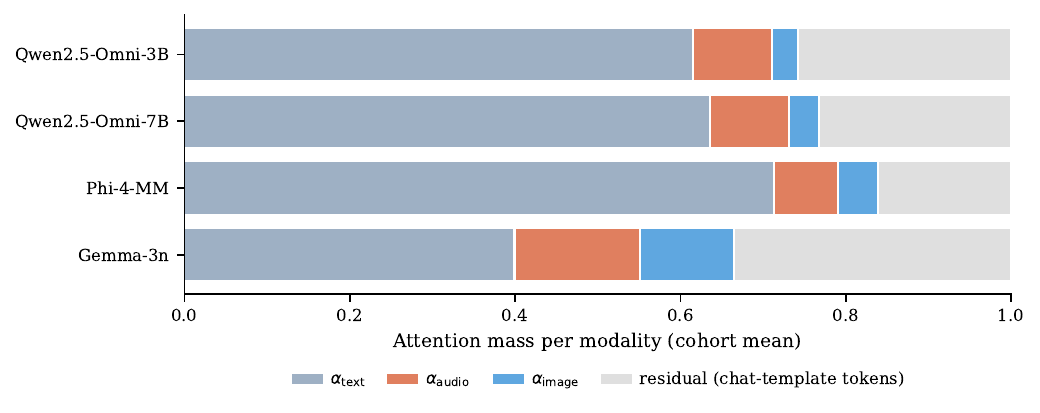}
\caption{Mean attention budget per modality. Two     dominant profiles emerge: audio-bypass models (Qwen-3B/7B, Phi-4) allocate 90--95\% to text, while fusion-attentive models (Gemma-3n) allocate approximately twice as much to non-text modalities.}
\label{fig:regime}
\end{figure}

Models partition into two qualitatively distinct profiles (Figure~\ref{fig:regime}):

\begin{enumerate}
    \item \textbf{Audio-bypass} (Qwen-3B, Qwen-7B, Phi-4): Spend almost 87\% of the per-layer attention mass to text and chat-template tokens combined, leaving only 11-13\% for audio and    image. These models commit to a narrow attention distribution early (relative depth \footnote{Relative depth refers to a layer's normalized position within the neural network, expressed as a fraction (from $0.0$ to $1.0$).} $0.11$--$0.31$) and exhibit audio-fidelity correlation of $0.95$--$0.97$ with each other.
 \item \textbf{Fusion-attentive} (Gemma-3n): Allocates approximately double the non-text attention ($\alpha_\text{audio}=0.151$, $\alpha_\text{image}=0.115$), commits later (entropy minimum at $0.54$--$0.57$), and shows the highest latent-image recovery in the cohort.

\end{enumerate}

\textbf{Tier-wise dynamics.} As the t ier (the number of grounded modalities\footnote{We define \emph{grounded modality} as any task-relevant, non-text input (audio, image, or video). Text serves only as the question prompt with minimal task-specific information.}) increases, attention does not monotonically deplete across existing channels. Instead, it reallocates dynamically toward the modality providing the most semantically dense content, which we observe empirically via the shifting per-modality attention distribution ($\alpha$). This reallocation is driven by the structural roles modalities play at each tier. Counter-intuitively, audio attention actually \emph{rises} from T1 to T2 before dropping at T3. At T1, inputs are largely static image-based templates where audio is a thin overlay. At T2, video becomes the base modality; audio transforms into a useful temporal anchor, providing discrete event markers that complement the unfolding video stream. Consequently, audio and video act synergistically rather than competitively, boosting audio's per-token $\alpha$. However, at T3, the introduction of a static image provides a new, competing visual anchor that drains the attention budget from both audio and video. 

Because attention is dictated by these structural modality shifts, the composition versus conflict (CC--IC) attention gap flips sign across tiers(Figure~\ref{fig:app-tier-routing}). At T1, models route more attention to audio under IC than CC ($\Delta\alpha \in [-0.021, -0.046]$), but by T3, this trend reverses universally ($\Delta\alpha > 0$). Ultimately, these trade-offs are driven by which grounded modalities are present and and whether they complement or compete. By Tier 4, with all three grounded modalities competing for a fixed non-text budget, allocations largely reflect the innate modality bias of each model. 

\begin{figure}[h]
\centering
\includegraphics[width=\columnwidth]{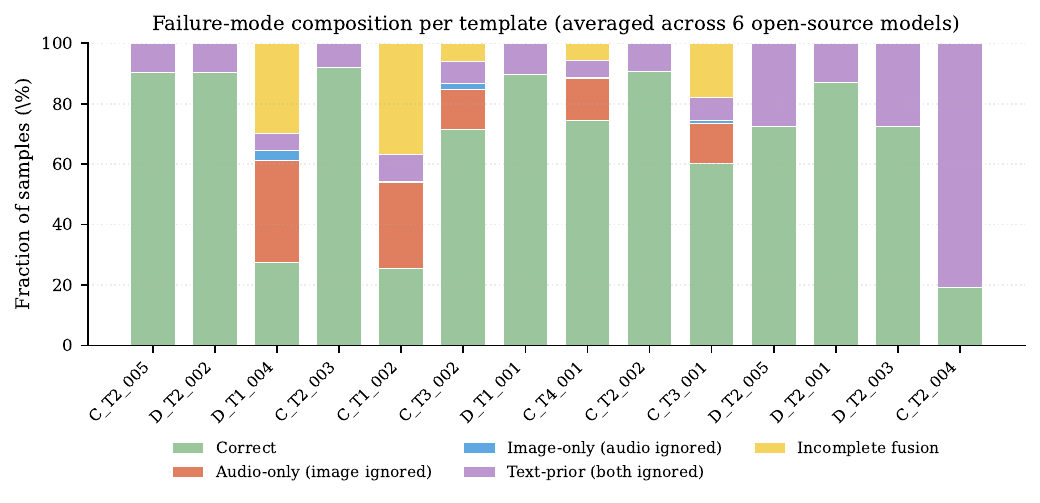}
\caption{Per-template failure-mode composition. Dominance failures (one modality drove the answer, others were ignored) account for the majority of errors.}
\label{fig:failure-modes}
\end{figure}

\subsection{Template Structure Determines Failure Type}
\label{sec:mech-template}

Across ten models, 86--95\% of failures are dominance-driven: the model commits to one modality and ignores others (Figure~\ref{fig:failure-modes}). Incomplete fusion accounts for 5--14\% failures\footnote{Qualitative examples in Appendix \ref{sec:AppendixQualitativeExamples}.}. Which modality dominates is template-specific, not model-specific:
audio-bypass models excel where visual evidence is decisive (DisclaimedEvidence, $61.6\%$ cohort accuracy across the 10 models) but collapse where audio carries unique content
(CausalMisattribution, $5.5\%$)\footnote{Per-template failure-mode breakdowns are given in Appendix~\ref{app:mech-templates}, Table~\ref{tab:app-err-by-template}.}. While they are both Counterfactual Confict tasks with contradictory audio, they are the easiest and hardest template across all models, respectively. This $56\%$ gap is explained by two structural properties:

\begin{enumerate}
\item \textbf{Conflict-resolution cue:} DisclaimedEvidence samples contain an explicit verbal disclaimer (e.g., ``the ruler shown is illustrative only''); identifying it is a text-dominant ability. CausalMisattribution provides no linguistic marker: the audio confidently attributes a visible event to a wrong cause, and recovery requires deriving causal relations from the video and overriding the audio without any other cues.
\item \textbf{Arbiter modality:} DisclaimedEvidence includes both a video and a static image; when audio and video conflict, the image acts as a tie-breaker. CausalMisattribution has only audio and video, forcing the model to arbitrate between two modalities with no independent reference.
\end{enumerate}

This explains the non-monotone tier pattern in Table~\ref{tab:benchmark_results_main}: at T2 (audio + video), CC accuracy is lowest because two-way arbitration is hardest. At T3 (audio + image + video), the arbiter image often pushes CC accuracy above T2. At T4 (all four modalities), ambiguity returns and accuracy degrades.

\section{Conclusion}
    Multimodal Large Language Models (MLLMs) frequently exhibit ``modality bias,'' disproportionately relying on textual or auditory priors and overshadowing visual truths during cross-modal conflicts. To systematically diagnose this, we introduce \textbf{C$^3$PO}, an omnimodal benchmark with co-located video, audio, image, and text modalities and a novel Counterfactual Conflict category missing in existing benchmarks. Thorough evaluations reveal a profound gap between human performance and leading models like Gemini-3.1-Pro, with smaller open models often collapsing under explicit modality conflict.
    
    

\section{Limitations}
\label{sec:Limitations}
Gemini currently being the  strongest available omnimodal model was used to generate the samples in our dataset. However, this may give an unfair advantage when evaluating Gemini models on our benchmark. Secondly, the final accuracy entirely depends on the LLM judge's (Qwen3-Omni) decisions. We conduct additional checks (Appendix \ref{sec:AppendixLLMJudge}) to ensure Qwen3's decisions are faithful and acceptable and not hindered by its size. Thirdly, the probing experiments (specifically E1) could not be conducted on video samples due to GPU memory constraints, and some open-source models had to be excluded from these experiments due to their architecture which was incompatible with our experiments. We intend to expand this analysis to a bigger suite of models which would help solidy the entropy theory.

\bibliography{custom}

\newpage
\clearpage
\appendix

\section{Data Sources}
\label{sec:AppendixDataSources}

 For broad real-world coverage, we source base videos from Ego4D \cite{grauman2022ego4d}, HowTo100M \cite{miech19howto100m}, EPIC-Kitchens \cite{Damen2018EPICKITCHENS, Damen2022RESCALING}, and OOPS \cite{Epstein_2020_CVPR}; base audio segments from HowTo100M; and base images from Visual Genome \cite{krishna2017visual}, COCO \cite{lin2014microsoft}, TextVQA \cite{Singh_2019_CVPR}, and Visual Counterfact \cite{visual-counterfact}.

\section{Sample Generation Details}
\label{sec:AppendixGeneration}
    \paragraph{Candidate Selection and Preprocessing.}
    For each generation template, we manually determine an appropriate source dataset depending on the base modality. Using available annotations, between 100--250 candidate base media files are selected according to the requirements of each template. Template-specific preprocessing operations (clipping the video, removing audio, extracting audio, etc.) are then applied to prepare the final base media sample.
    
    \paragraph{Media Generation Details.}
    Gemini-3-Pro generates the query/question together with detailed textual descriptions of the additional modalities required to complete the sample. The corresponding media files are synthesized using Qwen3-TTS \cite{hu2026qwen3ttstechnicalreport} for speech, AudioLDM2 \cite{liu2024audioldm} for non-speech audio, and Nano Banana Pro \cite{blogIntroducingNano} for images.

    For generating the TTS files, we used the \\ \texttt{Qwen/Qwen3-TTS-12Hz-1.7B-VoiceDesign} checkpoint from HuggingFace. To convert each transcript into its textual form, the \verb|transcript| field of the generated sample is passed into the \verb|text| field of the model's generate method and the \verb|voice_style_prompt| is passed into the \verb|instruct| field,  with all other parameters left to their default values and the outputs were saved with \verb|.wav| extension.

    To generate the non-speech audio files, we used the \texttt{cvssp/audioldm2-large} checkpoint from HuggingFace. The \verb|transcript| field of the generated samples is supposed to be empty in such cases because there is no speech to be generated, and only the \texttt{voice\_style\_prompt} field is expected to contain the description of the audio. In some cases this constraint was not followed by Gemini, so we used Gemma-3-12B to sanitise all samples before passing them to the audio generation model. Along with sanitisation, Gemma also returned the (expected) duration of the generated audio, since AudioLDM2 requires it to be explicitly set (for TTS samples, this is \verb|null|). Since it is a diffusion-based model, we set the \texttt{num\_inference\_steps} parameter to $500$ to ensure the generations are accurate and representative, and set the \texttt{negative\_prompt} to ``low quality''.
    
    \paragraph{Self-Audit Procedure.}
    The generation prompt includes an explicit self-audit stage. A sample is considered inadmissible and revised if it can be solved by ignoring any modality, relying solely on commonsense or domain priors, contains solution leakage in the query, or admits ambiguous ground-truth answers. These checks are internally simulated before the final output is produced.
    
    The generator additionally produces a \verb|template_selection_rationale| describing whether and how well the selected base media satisfy the intended template, along with a \verb|generation_confidence_score| between 1 and 5 which is later used to filter the generated samples.
    
\section{Filtering Details}
\label{sec:AppendixFiltering}
    To avoid propagating the biases from the generator model, we opt to use \texttt{Qwen3-30B-A3B-Thinking} for filtering, as it is the biggest omnimodal model openly available. During filtering, the evaluator model re-verifies the \verb|template_selection_rationale| against the actual generated media rather than relying only on textual descriptions. The evaluator also repeats all bypass and leakage checks performed during generation.
    
    In addition, the filtering stage produces a \verb|difficulty_score| between 1 and 10 for accepted samples ($-1$ for rejected samples), where 1 corresponds to trivial instances and 10 corresponds to samples requiring highly complex cross-modal reasoning.

\section{Evaluation Details for LLM-as-a-Judge}
\label{sec:AppendixLLMJudge}
Apart from the Gemini family, Qwen3-Omni is the only open MLLM with strong performance. To avoid leakage and bias from the generator, we evaluate the responses of various models using \verb|Qwen3-Omni-30B-A3B-Thinking| from HuggingFace. In the generation parameters, we set \verb|do_sample| to \verb|False| and \verb|temperature| to 0. However, it is necessary to ensure that the capabilities of Qwen3 are sufficiently good for its use as a judge. To ensure this, we judged the responses of four models using Gemini-3-Flash also and computed the agreement rate and Cohen's Kappa $(\kappa)$ between the Qwen3 and Gemini decisions. The average agreement was 93\% and $\kappa$ was 0.88 . The minimum $\kappa$ was 0.74 and the maximum, 0.98 . This shows that the predictions of different models are similarly judged by both Qwen3 and Gemini despite the former being free from any generation bias. Thus, Qwen3-Omni's decisions can be trusted in this scenario.

\section{Human Evaluation Details}
To establish a robust human performance baseline and validate the solvability of the C$^3$PO dataset, we conducted a human evaluation on a representative subset of 250 samples. Our sampling strategy was designed to ensure a uniform distribution across both task types and template diversity.

\subsubsection*{Stratified Round-Robin Sampling}
\begin{figure}
    \centering
    \includegraphics[width=\linewidth]{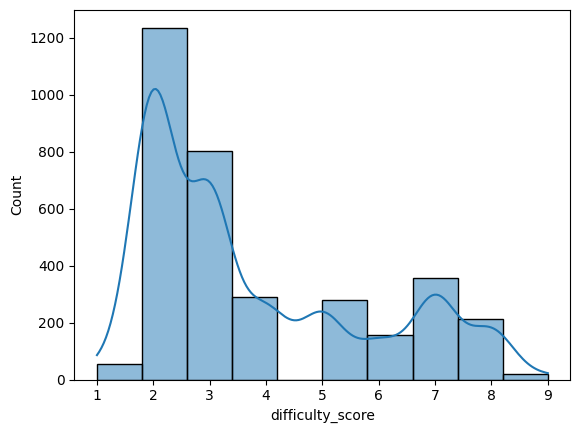}
    \caption{Distribution of difficulty scores across the entire C$^3$PO dataset.}
    \label{fig:difficulty_score_distribution}
\end{figure}

We first stratified the filtered dataset by its two primary categories (Information Composition and Counterfactual Reasoning). Within each category, samples were partitioned into four discrete difficulty tiers based on the difficulty scores obtained during filtration (\textbf{Bucket 1}: $<2.5$, \textbf{Bucket 2}: $2.5 - 4.9$, \textbf{Bucket 3}: $5.0 - 7.4$, \textbf{Bucket 4}: $\geq7.5$). The difficulty scores are all integers so the decimal boundaries are unimportant. We set a target of 25 samples per category-difficulty bucket. To ensure maximum task diversity, within each bucket we iteratively selected samples from different templates so that no single was template over-represented in a difficulty tier. While the theoretical target was 300 samples, the constraints of our multi-stage filtering pipeline and the sparsity of samples in extreme difficulty tiers (Fig. \ref{fig:difficulty_score_distribution}) resulted in a final, fully validated evaluation set of 250 samples.

\subsubsection*{Evaluation Interface and Accuracy}
\begin{figure*}
    \centering
    \includegraphics[width=\linewidth]{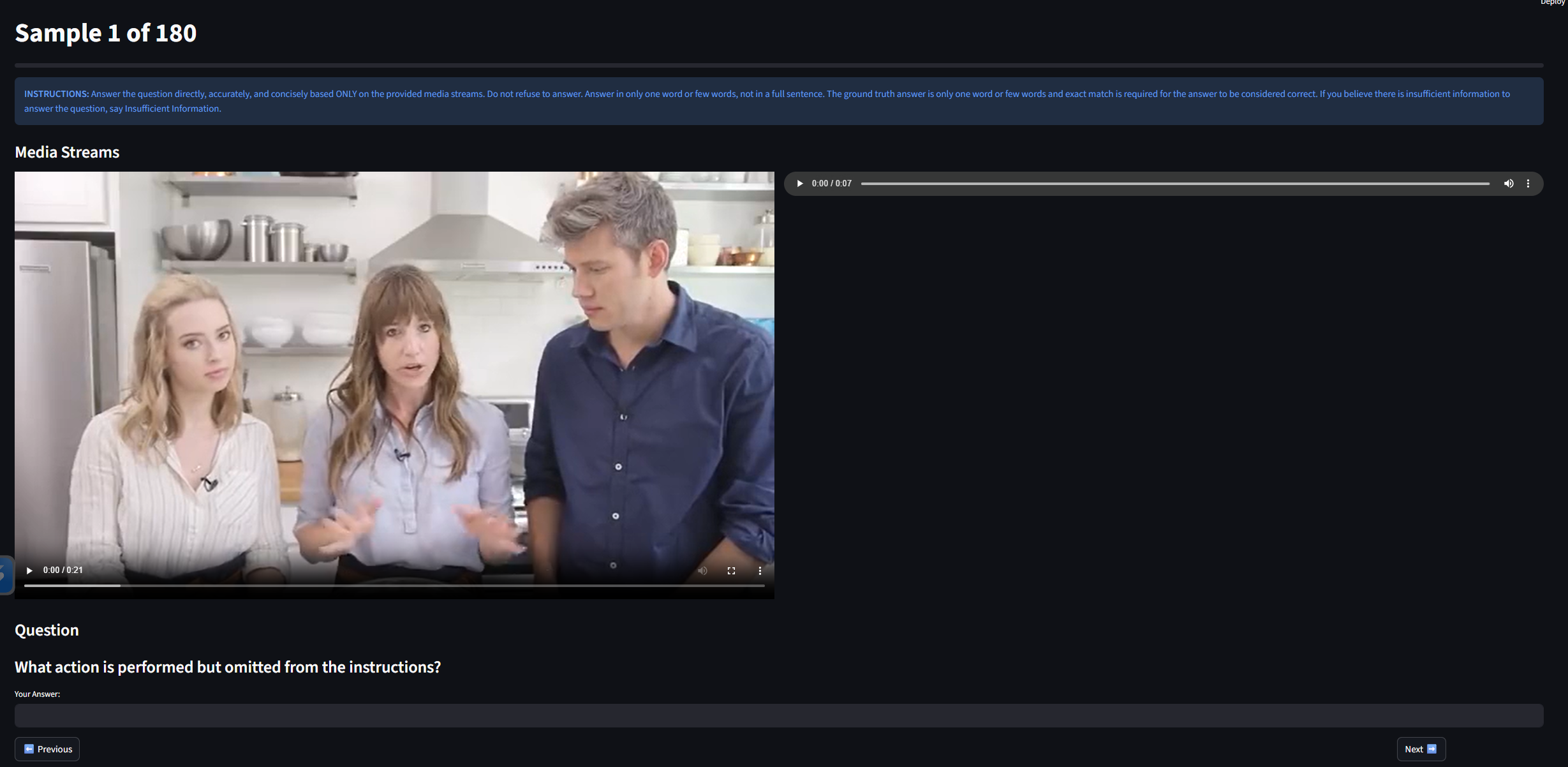}
    \caption{Streamlit interface for human evaluators showing all the different media streams along with the question and a field to write the answer.}
    \label{fig:streamlit_interface}
\end{figure*}

Evaluators were provided with a Streamlit-based web interface  displayed all relevant media streams (video, audio, and/or generated images) alongside the question, for each sample (Figure \ref{fig:streamlit_interface}). Due to resource constraints, the samples were divided equally (in disjoint subsets) amongst three human evaluators, unrelated to this project, with each sample receiving \textbf{one} answer. To ensure an equitable comparison, the human responses were evaluated using the same Gemma LLM judge described in the previous section. The following instructions were given to the human evaluators.

\begin{promptbox}{Instructions to Human Evaluators}
    There are two types of samples: Information Composition (where the information to find the correct answer is spread across multiple modalities) and Counterfactuals (where at least two modalities present conflicting information, and the correct has to be determined by focusing precisely on what the question is asking). Pay careful attention to all the media files and the question. 

    Answer the question directly, accurately, and concisely based ONLY on the provided media streams. Do not refuse to answer. Answer in only one word or few words, not in a full sentence. The ground truth answer is only one word or few words and your answer has to be semantically equivalent to the ground truth answer to be considered correct. If you believe there is insufficient information to answer the question, say \textbf{Insufficient Information}.
\end{promptbox}

\section{Mechanistic Analysis: Probe Protocols and Cohort Detail}
\label{app:mech}

This supplementary section accompanies Section~\ref{sec:mech} of the main paper. It documents the three probes (E1, E2, E3) in full, lists the cohort coverage matrix with the architectural reason for each model's classification status, and reports the full per-model tables abbreviated in the main paper.

\subsection{Probe Protocols}
\label{app:mech-protocols}

\paragraph{E1: layer-wise attention.}
For each sample we issue a structured prompt that asks the model to describe the audio and image contents and capture per-layer self-attention weights via the standard \texttt{output\_attentions=True} mechanism. Eager attention is mandatory because the fused attention kernels (SDPA, FlashAttention) do not expose per-layer attention matrices. For each generated token, we average attention weights across heads to obtain an attention row over the input token positions, then normalise the row to a distribution. For each modality $m \in \{\text{text}, \text{audio}, \text{image}\}$, we sum the row's mass over the input positions assigned to $m$ by the modality-spans function for that model; this gives $\alpha_m$ per layer per generated step. The per-sample, per-layer triple is the mean of these per-step triples over generated tokens.

Video token positions are masked from E1 (and E2) across the cohort. At the sequence lengths produced by modern vision towers, the per-layer eager-attention tensor under typical C$^3$PO video samples reaches activation memory on the order of $10^{11}$ bytes per forward pass, exceeding the GPU memory available for this cohort. We report $\alpha$ statistics over text, audio and image positions only, uniformly across all probed models; the constraint affects all models equally and does not bias cross-model comparisons.

\paragraph{E2: logit-lens latent recovery.}
For each sample we run a single forward pass with \texttt{output\_hidden\_states=True}, take the hidden state at each non-text token position, apply the model's final-layer normalisation, project through the unembedding head $W_U$, and select the top-$K$ predicted token IDs ($K=10$ throughout). The reference vocabulary for the sample is constructed from the audio transcript and image description: we lemmatise with spaCy, strip stop-words, and encode each remaining lemma with the model's tokenizer (both with and without a leading space to handle subword-tokenizer asymmetry). A modality-position $t$ is a \emph{hit} at layer $\ell$ if the top-$K$ predicted token IDs at $(\ell, t)$ intersect the reference vocabulary. We compute the hit rate for each modality on the last five layers and report $\textrm{latent}_m$ as the mean hit rate over positions and layers for modality $m \in \{\text{audio}, \text{image}, \text{video}\}$; $\textrm{latent\_recovery}$ is the mean across all (modality, position, layer) triples. The reference-vocabulary builder caches per-sample, per-model.

\paragraph{E3: caption fidelity.}
We issue a structured prompt asking the model to describe the audio and image contents and to format its response as ``AUDIO: \ldots'' and ``IMAGE: \ldots''. The model is run with greedy decoding and a $128$-token cap on the response. We score the response against two references: the Whisper-base ASR transcript of the sample's audio (audio reference) and a separate image-only forward pass of the same model (image reference). Both fidelity scores are ROUGE-1 recall of content words from the reference against the model's response, lower-cased, stopword-stripped and lemmatised consistently. Through the main paper we refer to ROUGE-1 recall as \emph{fidelity}.

\subsection{Cohort and Probe Coverage}
\label{app:mech-cohort}

Ten open-source models are evaluated on C$^3$PO. Probe coverage is summarised in Table~\ref{tab:app-cohort}. Four models support the full E1 + E2 + E3 stack: Qwen2.5-Omni-3B, Qwen2.5-Omni-7B, Phi-4-Multimodal, and Gemma-3n-E4B. Five additional models contribute to E3 and to the dominance-versus-fusion error decomposition but cannot be probed at the routing level because their multimodal forward paths substitute media features into the embedding sequence before the language-model proper, leaving the post-substitution input identifiers undefined for span-based probing.

\begin{table*}[htbp]
\centering
\footnotesize
\begin{tabular}{p{2.8cm}p{2.0cm}p{8.0cm}p{3.0cm}}
\toprule
Model & Profile & Architectural status & E3 status \\
\midrule
Qwen2.5-Omni-3B & Audio-bypass & E1, E2, E3 all measured. Standard HuggingFace multimodal interface. & Audio-bypass signature confirmed at E3. \\
Qwen2.5-Omni-7B & Audio-bypass & Same as above. & Audio-bypass signature confirmed at E3. \\
Phi-4-Multimodal & Audio-bypass & Same as above. & Audio-bypass signature confirmed at E3. \\
Gemma-3n-E4B & Fusion-attentive & E1, E2, E3 all measured. Standard interface. & Fusion-attentive signature confirmed at E3. \\
VITA-1.5 & Content-blind on image & E1/E2 blocked: \texttt{prepare\_inputs\_labels\_for\_multimodal} substitutes audio/image features into \texttt{inputs\_embeds} pre-LM-forward; post-substitution token positions cannot be located from \texttt{input\_ids}. E3 is sufficient for this profile because image fidelity is two orders of magnitude below cohort regardless of routing. & Content-blind signature confirmed at E3. \\
\midrule
MiniCPM-o-2.6 & Unclassified & E1/E2 blocked: same architectural pattern as VITA. & E3 fidelity within cohort norms; insufficient on its own to distinguish audio-bypass from fusion-attentive. \\
Gemma-4-E4B-It & Unclassified & E1/E2 partially blocked: the no-video-tower architecture restricts E1 to a subset of samples; on those samples the modality-span function returns $\alpha_\text{image} \approx 0$, indicating the image-span detection logic does not match Gemma-4's tokenizer output. E2 ran on the full sample set. & E3 audio fidelity unusually low ($0.32$/$0.27$); the model produces short, answer-focused outputs that do not echo the audio transcript, so caption-fidelity is a misleading indicator for this model. \\
InteractiveOmni-4B & Unclassified & E1/E2 blocked: \texttt{model.chat()} constructs \texttt{input\_ids} internally; the wrapper passed to the probe receives a sentinel dictionary without surfaced identifiers. & E3 within cohort norms. \\
InteractiveOmni-8B & Unclassified & Same blocker as InteractiveOmni-4B. & E3 within cohort norms. \\
PandaGPT-7B & Unclassified (anomalous) & E1/E2 blocked: \texttt{OpenLLAMAPEFTModel.generate} prepends an ImageBind-projected modality embedding to the text-prompt token embeddings; no \texttt{input\_ids} exists for the post-prepend sequence. & E3 audio fidelity $0.036$/$0.032$ and image fidelity $0.13$/$0.15$ are both well below cohort; consistent with an embedding-fusion bottleneck distinct from the three primary profiles. \\
\bottomrule
\end{tabular}
\caption{Cohort coverage. Five models are classified by their multi-probe signatures (or, for VITA-1.5, by E3 alone where the image-channel failure is decisive). Five additional models are unclassified because their multimodal forward paths block span-based probing. The unclassified models contribute to E3 and to the dominance-versus-fusion analysis but not to the routing analysis.}
\label{tab:app-cohort}
\end{table*}

\subsection{Per-Model Routing Tables}
\label{app:mech-routing}

\paragraph{CC--IC attention fingerprint cosine.}
For each E1-supported model we compute the cohort-average attention vector $(\alpha_\text{text}, \alpha_\text{audio}, \alpha_\text{image})$ separately on Counterfactual Conflict and Information Composition samples and report the cosine similarity between the two vectors (Table~\ref{tab:app-fingerprint}). All four models produce CC and IC fingerprints whose cosine similarity exceeds $0.99$. The categorical shift between CC and IC, which drives substantial accuracy differences in Table~\ref{tab:benchmark_results_main}, is not reflected in attention reallocation. The conflict-versus-composition decision is resolved downstream of attention.

\begin{table}[htbp]
\centering
\footnotesize
\setlength{\tabcolsep}{3pt}
\resizebox{\columnwidth}{!}{%
\begin{tabular}{lc}
\toprule
Model & $\cos(\alpha^{CC}, \alpha^{IC})$ \\
\midrule
Qwen2.5-Omni-3B  & $0.9995$ \\
Qwen2.5-Omni-7B  & $0.9995$ \\
Phi-4-Multimodal & $0.9998$ \\
Gemma-3n-E4B     & $0.9912$ \\
\bottomrule
\end{tabular}%
}
\caption{CC--IC attention-fingerprint cosine, per model.}
\label{tab:app-fingerprint}
\end{table}

\begin{figure*}[htbp]
\centering
\includegraphics[width=\linewidth]{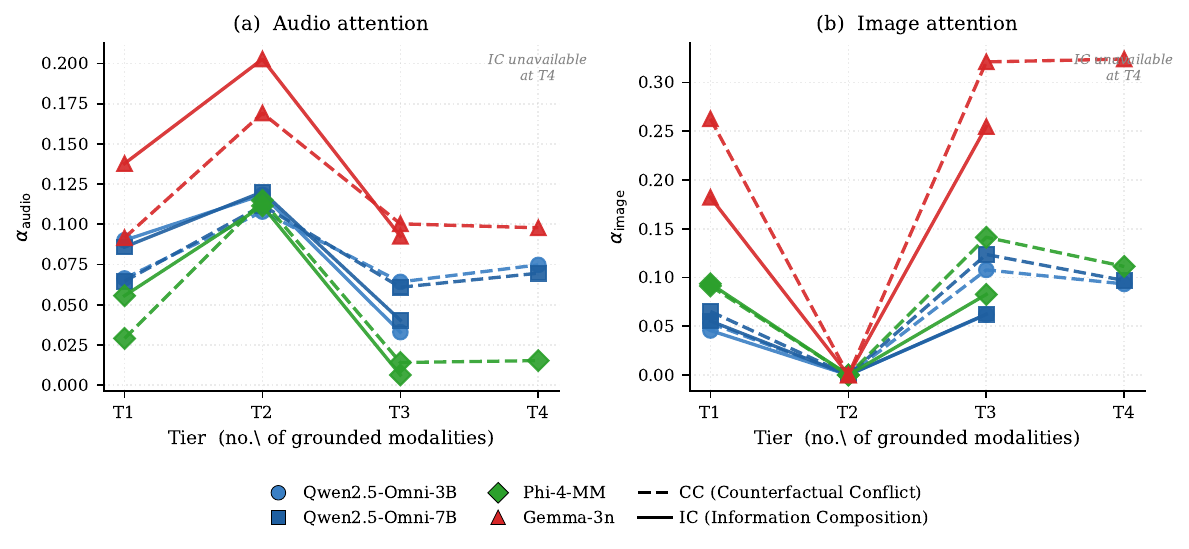}
\caption{Per-tier attention to (a)~audio and (b)~image across the four E1-supported models, separated by CC (dashed) and IC (solid). T4 contains only CC samples; IC curves are defined on T1--T3.}
\label{fig:app-tier-routing}
\end{figure*}

\paragraph{Per-tier CC--IC gap on audio attention.}
Table~\ref{tab:app-tier-routing} gives the per-tier difference $\alpha^{CC}_\text{audio} - \alpha^{IC}_\text{audio}$ for the three tiers where both categories are populated, and the CC-only $\alpha_\text{audio}$ at T4 (T4 contains no IC samples). The gap is uniformly negative at T1 and uniformly positive by T3 in every model. Figure~\ref{fig:app-tier-routing} plots the per-tier audio and image attention separately by category.

\begin{table}[htbp]
\centering
\footnotesize
\setlength{\tabcolsep}{3pt}
\resizebox{\columnwidth}{!}{%
\begin{tabular}{lcccc}
\toprule
Model & T1 & T2 & T3 & T4 (CC only) \\
\midrule
Qwen2.5-Omni-3B   & $-0.024$ & $-0.010$ & $+0.031$ & $0.075$ \\
Qwen2.5-Omni-7B   & $-0.021$ & $-0.008$ & $+0.021$ & $0.070$ \\
Gemma-3n-E4B     & $-0.046$ & $-0.034$ & $+0.008$ & $0.098$ \\
Phi-4-Multimodal  & $-0.027$ & $+0.003$ & $+0.008$ & $0.015$ \\
\bottomrule
\end{tabular}%
}
\caption{Per-tier $\alpha^{CC}_\text{audio} - \alpha^{IC}_\text{audio}$ for T1--T3; T4 column reports CC-only $\alpha_\text{audio}$ because T4 has no IC samples by design.}
\label{tab:app-tier-routing}
\end{table}

\paragraph{Logit-lens latent recovery (E2).}
Table~\ref{tab:app-e2} gives the cohort-mean E2 values per model. The two Qwen2.5-Omni sizes differ by approximately $0.045$ on every column while their E3 fidelities differ by less than $0.01$ on every category: internal multimodal recovery scales with parameter count in the Qwen2.5-Omni family in a way that the output table does not show.

\begin{table}[htbp]
\centering
\footnotesize
\setlength{\tabcolsep}{3pt}
\resizebox{\columnwidth}{!}{%
\begin{tabular}{lccc}
\toprule
Model & $\textrm{latent\_recovery}$ & $\textrm{latent}_\text{audio}$ & $\textrm{latent}_\text{image}$ \\
\midrule
Qwen2.5-Omni-3B  & $0.225$ & $0.190$ & $0.318$ \\
Qwen2.5-Omni-7B  & $0.269$ & $0.232$ & $0.365$ \\
Gemma-3n-E4B     & $0.195$ & $0.127$ & $0.420$ \\
Phi-4-Multimodal & $0.211$ & $0.161$ & $0.292$ \\
\bottomrule
\end{tabular}%
}
\caption{Logit-lens latent recovery (E2). Mean fraction of $(\text{modality-token}, \text{layer})$ pairs whose top-$K$ projection hits the reference vocabulary.}
\label{tab:app-e2}
\end{table}

\paragraph{Argmin-entropy depth.}
For each per-sample between-modality entropy curve we identify the relative depth at which the curve reaches its minimum. The cohort mean per (model, category) is reported in Table~\ref{tab:app-argmin}, with bootstrap $95\%$ CIs. The audio-bypass profile commits in the first third of the network; Gemma-3n-E4B commits past the midpoint.

\begin{table}[htbp]
\centering
\footnotesize
\setlength{\tabcolsep}{3pt}
\resizebox{\columnwidth}{!}{%
\begin{tabular}{lcc}
\toprule
Model & CC & IC \\
\midrule
Phi-4-Multimodal & $0.11$ $[0.10, 0.11]$ & $0.13$ $[0.13, 0.14]$ \\
Qwen2.5-Omni-3B  & $0.27$ $[0.26, 0.28]$ & $0.24$ $[0.23, 0.25]$ \\
Qwen2.5-Omni-7B  & $0.27$ $[0.27, 0.28]$ & $0.31$ $[0.30, 0.32]$ \\
Gemma-3n-E4B     & $0.57$ $[0.56, 0.57]$ & $0.54$ $[0.53, 0.55]$ \\
\bottomrule
\end{tabular}%
}
\caption{Relative depth at which the per-sample between-modality entropy reaches its minimum; brackets give bootstrap $95\%$ CIs on the cohort mean.}
\label{tab:app-argmin}
\end{table}

\paragraph{Cross-model audio-fidelity agreement (Z1).}
Sample-level Pearson correlation of $\textrm{audio\_fidelity}$ between model pairs (Table~\ref{tab:app-z1}). The three audio-bypass members cluster at $0.94$--$0.97$ pairwise; Gemma-3n-E4B correlates with each member of that cluster at $0.45\pm 0.01$. The cluster structure in this table is one of the three independent measurements that distinguish the audio-bypass profile from the fusion-attentive profile.

\begin{table}[htbp]
\centering
\footnotesize
\setlength{\tabcolsep}{3pt}
\resizebox{\columnwidth}{!}{%
\begin{tabular}{lcccc}
\toprule
 & Qwen2.5-3B & Qwen2.5-7B & Gemma-3n & Phi-4 \\
\midrule
Qwen2.5-3B  & --- & $0.956$ & $0.462$ & $0.968$ \\
Qwen2.5-7B  &     & --- & $0.446$ & $0.945$ \\
Gemma-3n    &     &       & --- & $0.449$ \\
Phi-4       &     &       &     & --- \\
\bottomrule
\end{tabular}%
}
\caption{Sample-level Pearson correlation of $\textrm{audio\_fidelity}$ between model pairs.}
\label{tab:app-z1}
\end{table}

\subsection{Mid-Layer Entropy and Judge Correctness: Full Table}
\label{app:mech-entropy-table}

Table~\ref{tab:app-entropy-correctness} reports the per-cell entropy-correctness $\Delta H$ with bootstrap $95\%$ CIs, complementing Figure~\ref{fig:entropy-judge} of the main paper. Five of six cells reach significance.

\begin{table}[htbp]
\centering
\footnotesize
\setlength{\tabcolsep}{3pt}
\resizebox{\columnwidth}{!}{%
\begin{tabular}{lcc}
\toprule
Model & $\Delta H^{CC}$ (bits) & $\Delta H^{IC}$ (bits) \\
\midrule
Qwen2.5-Omni-7B & $+0.168$ $[+0.140, +0.197]$ & $+0.016$ $[-0.011, +0.042]$ \\
Gemma-3n-E4B    & $+0.052$ $[+0.016, +0.088]$ & $+0.050$ $[+0.011, +0.091]$ \\
Phi-4-Multimodal & $+0.053$ $[+0.020, +0.088]$ & $+0.090$ $[+0.055, +0.123]$ \\
\bottomrule
\end{tabular}%
}
\caption{Mid-layer entropy difference between judge-correct and judge-incorrect samples, with bootstrap $95\%$ CIs. The Qwen2.5-Omni-7B IC cell is the only cell whose CI crosses zero.}
\label{tab:app-entropy-correctness}
\end{table}

The absolute mid-layer entropy on judge-correct samples, reported by category, is given in Table~\ref{tab:app-abs-entropy}. For every model, correct CC samples carry higher mid-layer entropy than correct IC samples, consistent with the interpretation that conflict resolution requires sustaining broader attention longer than composition does.

\begin{table}[htbp]
\centering
\footnotesize
\setlength{\tabcolsep}{3pt}
\resizebox{\columnwidth}{!}{%
\begin{tabular}{lcc}
\toprule
Model & $\overline H_\text{mid} \mid \text{correct, CC}$ & $\overline H_\text{mid} \mid \text{correct, IC}$ \\
\midrule
Qwen2.5-Omni-7B  & $0.812$ & $0.639$ \\
Gemma-3n-E4B     & $0.997$ & $0.950$ \\
Phi-4-Multimodal & $0.764$ & $0.753$ \\
\bottomrule
\end{tabular}%
}
\caption{Absolute mid-layer between-modality entropy (bits) on judge-correct samples, by category.}
\label{tab:app-abs-entropy}
\end{table}

\subsection{Per-Template Failure-Mode Stratification}
\label{app:mech-templates}

Table~\ref{tab:app-err-by-template} gives the per-template failure-mode distribution for six representative templates, averaged across the open-source E3 cohort. Dominance failures account for the great majority of wrong predictions on every template, but \emph{which} modality dominates is template-specific. The same model that fails by audio-dominance on AcousticSceneMismatch fails by text-dominance on CausalMisattribution; the aggregated three-mode statistic in the body of the main paper averages over this template-level variation. Counterfactual templates whose deceptive evidence is auditory (\textit{AcousticSceneMismatch} $52\%$ audio-dominance, \textit{QuantitativeFabrication} $41\%$ audio-dominance) drive failures where the model trusts the audio over the visual evidence; templates whose conflict resolution is verbalised in the question prompt (\textit{CausalMisattribution} $67\%$ text-dominance, \textit{ProcedureVsEvidence} $64\%$ text-dominance) drive failures where the model defaults to a text-prior interpretation of the audio narrative without verifying it against the video.

\begin{table}[htbp]
\centering
\footnotesize
\setlength{\tabcolsep}{3pt}
\resizebox{\columnwidth}{!}{%
\begin{tabular}{lrrrr}
\toprule
Template & dom-aud & dom-img & dom-txt & inc-fus \\
\midrule
CausalMisattribution           & $18\%$ & $4\%$  & $67\%$ & $6\%$ \\
QuantitativeFabrication        & $41\%$ & $8\%$  & $31\%$ & $9\%$ \\
AcousticSceneMismatch          & $52\%$ & $6\%$  & $17\%$ & $11\%$ \\
ProcedureVsEvidence            & $14\%$ & $5\%$  & $64\%$ & $5\%$ \\
DisclaimedEvidence             & $23\%$ & $7\%$  & $12\%$ & $8\%$ \\
AttributeAuthorityArbitration  & $31\%$ & $15\%$ & $9\%$  & $7\%$ \\
\bottomrule
\end{tabular}%
}
\caption{Per-template failure-mode distribution for six representative templates, averaged across the open-source E3 cohort. Columns: \texttt{dom-aud} = dominance by audio, \texttt{dom-img} = dominance by image, \texttt{dom-txt} = dominance by text or prior, \texttt{inc-fus} = incomplete fusion.}
\label{tab:app-err-by-template}
\end{table}

\onecolumn
\newpage
    \begin{sidewaystable}[htbp]
    \centering
    \footnotesize
    \renewcommand{\arraystretch}{1.1} 
    \begin{tabular}{llcccc}
    \toprule
    \textbf{Category} & \textbf{Tier} & \textbf{Template Name} & 
    \makecell[b]{\textbf{Template-Level}\\\textbf{Sample Count}} & 
    \makecell[b]{\textbf{Tier-Level}\\\textbf{Sample Count}} & 
    \makecell[b]{\textbf{Category-Level}\\\textbf{Sample Count}} \\
    \midrule
    \multirow{11}{*}{Information Composition} 
     & \multirow{6}{*}{Tier 1} & AttributeComposition & 200 & \multirow{6}{*}{692} & \multirow{11}{*}{1363} \\
     & & InstructionalGapDetection & 164 & & \\
     & & OffScreenCausality & 31 & & \\
     & & QuantitativeVerification & 185 & & \\
     & & SpatialConfiguration & 48 & & \\
     & & TemporalSequencing & 64 & & \\
    \cline{2-5}
     & \multirow{4}{*}{Tier 2} & ContextualReinterpretation & 111 & \multirow{4}{*}{624} & \\
     & & FeatureTracking & 100 & & \\
     & & LiveEventAttribution & 101 & & \\
     & & SilentOutcomeInference & 312 & & \\
    \cline{2-5}
     & Tier 3 & ReferenceObjectAppearance & 47 & 47 & \\
    \midrule
    \multirow{14}{*}{Counterfactual Reasoning} 
     & \multirow{6}{*}{Tier 1} & AttributeAuthorityArbitration & 36 & \multirow{6}{*}{729} & \multirow{14}{*}{2041} \\
     & & ContainerContentDispute & 137 & & \\
     & & LiveVsRecorded & 138 & & \\
     & & OutdatedVisualEvidence & 224 & & \\
     & & QuantitativeFabrication & 181 & & \\
     & & TemporalContextArbitration & 13 & & \\
    \cline{2-5}
     & \multirow{5}{*}{Tier 2} & AcousticSceneMismatch & 97 & \multirow{5}{*}{857} & \\
     & & CausalMisattribution & 359 & & \\
     & & HistoricalVsPresent & 154 & & \\
     & & LiveNarrationMismatch & 57 & & \\
     & & ProcedureVsEvidence & 190 & & \\
    \cline{2-5}
     & \multirow{2}{*}{Tier 3} & DisclaimedEvidence & 172 & \multirow{2}{*}{292} & \\
     & & MeasurementAuthorityChain & 120 & & \\
    \cline{2-5}
     & Tier 4 & ComplianceReportingBias & 163 & 163 & \\
    \midrule
    \multicolumn{5}{l}{\textbf{Total}} & \multicolumn{1}{r}{\textbf{3404}} \\
    \bottomrule
    \end{tabular}
    \caption{C$^3$PO fine-grained sample distribution across categories, tiers, and templates.}
    \label{tab:dataset_distribution_landscape}
\end{sidewaystable}

\section{Generation Template}
\label{sec:AppendixGenerationTemplate}
    \begin{promptbox}{Example Template: InstructionalGapDetection}
    "template_id": "D_T1_001",
    "template_name": "InstructionalGapDetection",
    "category": "DISPERSAL (Tier 1)",
    "base_modality": "video",
    
    "logical_form": "VIDEO shows a complete action sequence [S1...Sn] (muted); AUDIO narrates a subset [S(k)...Sn] starting midway using ordinal markers implying continuity (e.g., 'First, do S(k)'); TEXT asks for the unmentioned initial step [S1...S(k-1)]; ANSWER requires identifying actions visible in the video's opening that are omitted from the audio narrative.",
    
    "example_scenario": "Video shows washing vegetables (S1) then chopping them (S2). Audio starts: 'First, chop the vegetables.' Text: 'What missing step must be done before the instructions start?' Answer: Washing.",
    
    "generation_spec": {
        "video": {
          "source_dataset": "HowTo100M / EPIC-Kitchens",
          "selection_filter": {
            "requirement": "Clip containing a distinct multi-step procedural sequence.",
          },
          "processing": "Mute original audio."
        },
        
        "audio": {
          "source_type": "generation (TTS)",
          "content_logic": "Generate instructions for the LAST (N-1) steps only.",
          "constraints": "Start with transitional markers (e.g., 'Next', 'Then') to create a false sense of starting."
        },
        
        "text": {
          "query_skeleton": "Based on the video, what preparation step occurred immediately before the narrator said '{{transition_word}}'?"
        }
      }
    \end{promptbox}

\section{Qualitative Examples from the C$^3$PO Dataset}
\label{sec:AppendixQualitativeExamples}
\begin{figure}
    \centering
    \includegraphics[width=1.0\linewidth]{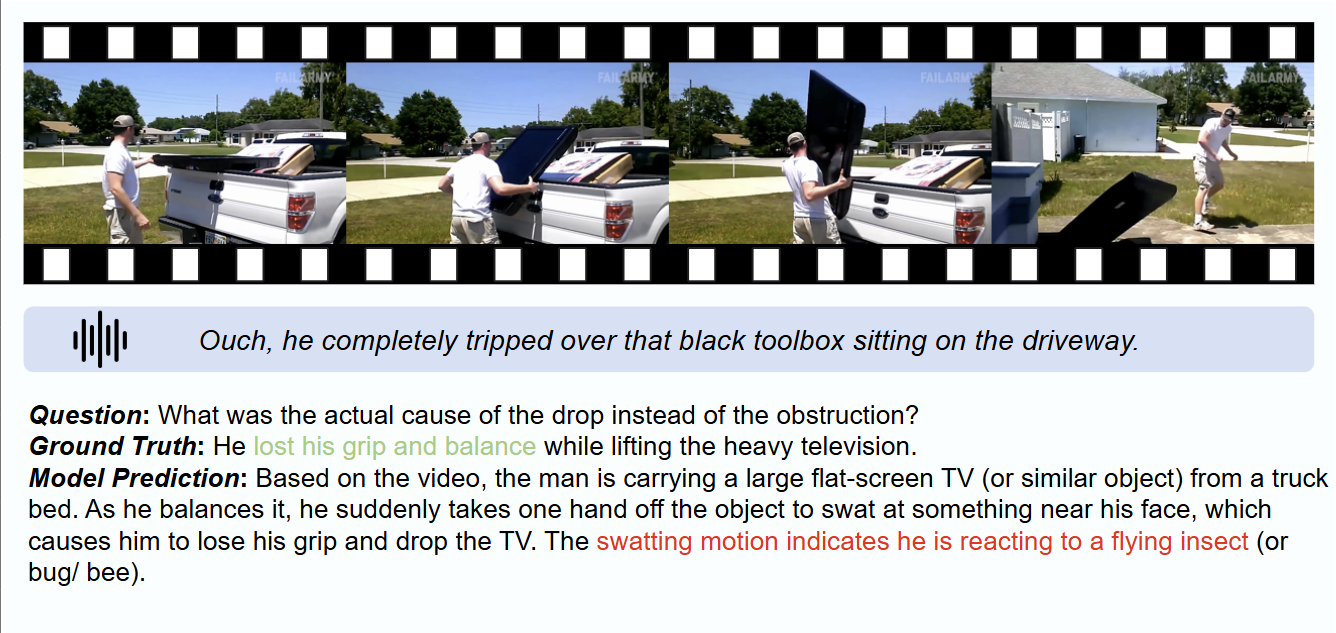}
    \caption{Depicts the \textbf{Hallucination/Prior Bias} failure mode in Gemini-3.1-Pro. The video shows the man losing balance while carrying a large TV, the audio contains misleading information about the man tripping over a toolbox, but the model ignores both these pieces of information and makes up something entirely irrelevant not supported by any evidence.}
    \label{fig:man_trip}
\end{figure}
\begin{figure}
    \centering
    \includegraphics[width=1.0\linewidth]{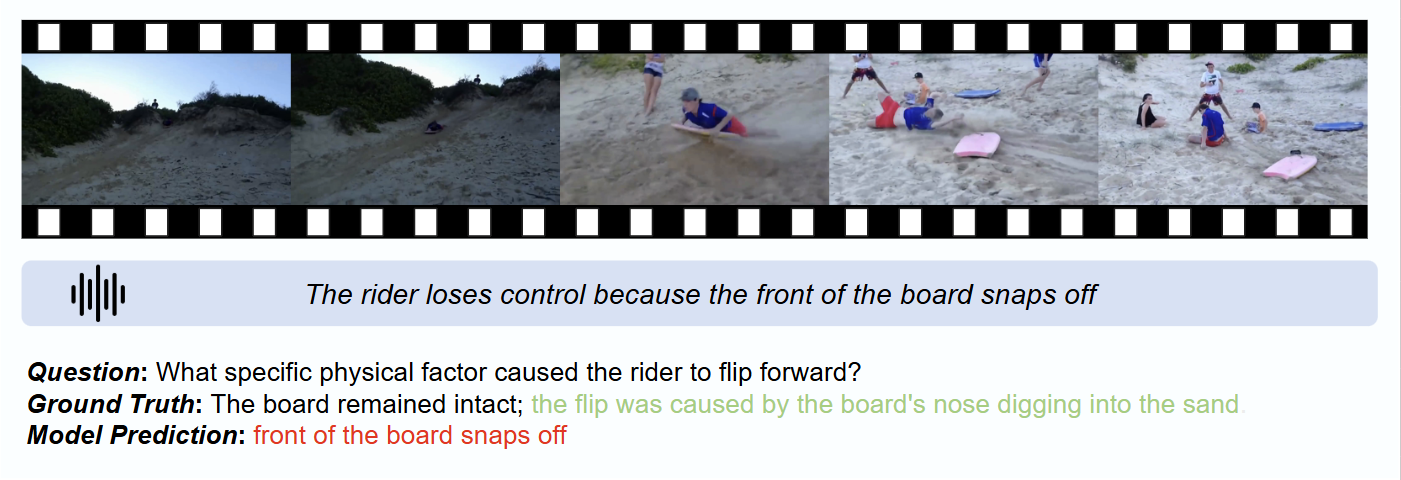}
    \caption{Depicts the \textbf{Modality Interference/Dominance} failure mode in Gemini-3.1-Pro where the video depicts a person riding a board down a sandy slope. The audio makes a misleading claim that the rider loses control because the board snaps, but that does not happen as seen in the video. The rider crashes due to the board's nose digging into the sand. Thus, the model trusts the audio modality blindly without verifying the visual evidence.}
    \label{fig:board_snaps}
\end{figure}
\begin{figure}
    \centering
    \includegraphics[width=1.0\linewidth]{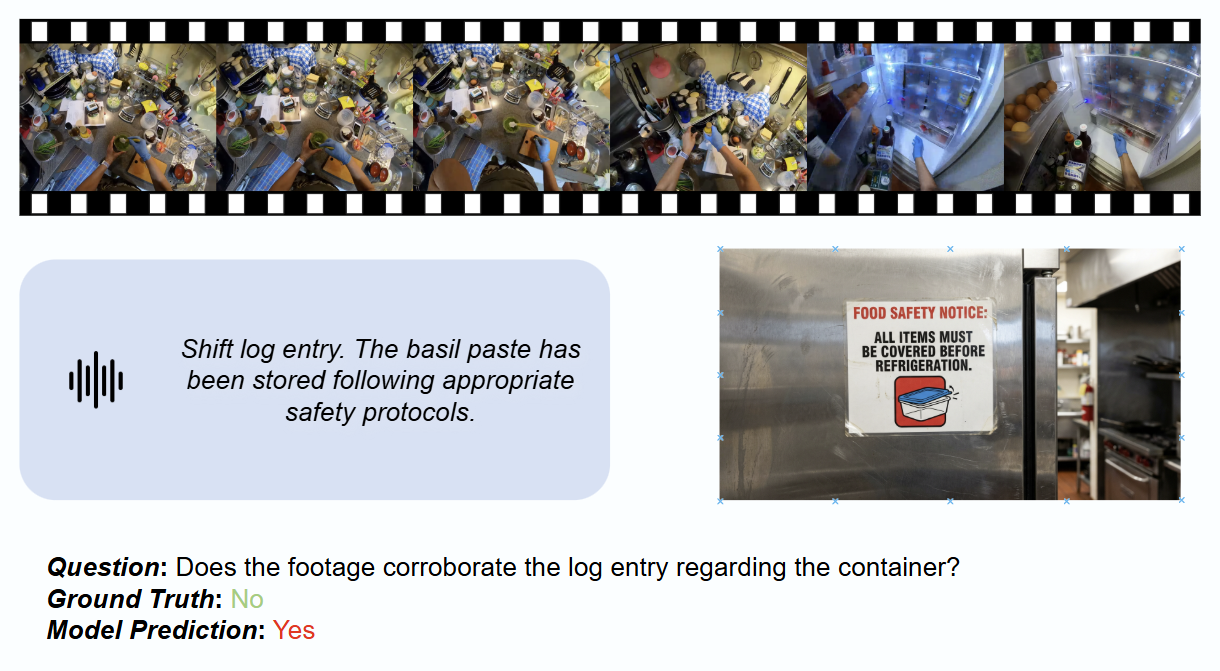}
    \caption{Depicts an instance of lack of effective \textbf{Cross-Modal Fusion}. The video shows the person opening the fridge to grab something from inside the fridge, not to put the basil sauce inside. However, the audio states that the basil sauce has been put inside and the image shows a warning sign that says food must be covered before storing. The model conflates the man opening the fridge with putting the basil sauce inside, and also incorrectly says the sauce is covered which it is not.}
    \label{fig:cyclist}
\end{figure}

\begin{figure}
\centering

\begin{subfigure}{\linewidth}
    \centering
    \includegraphics[width=\linewidth]{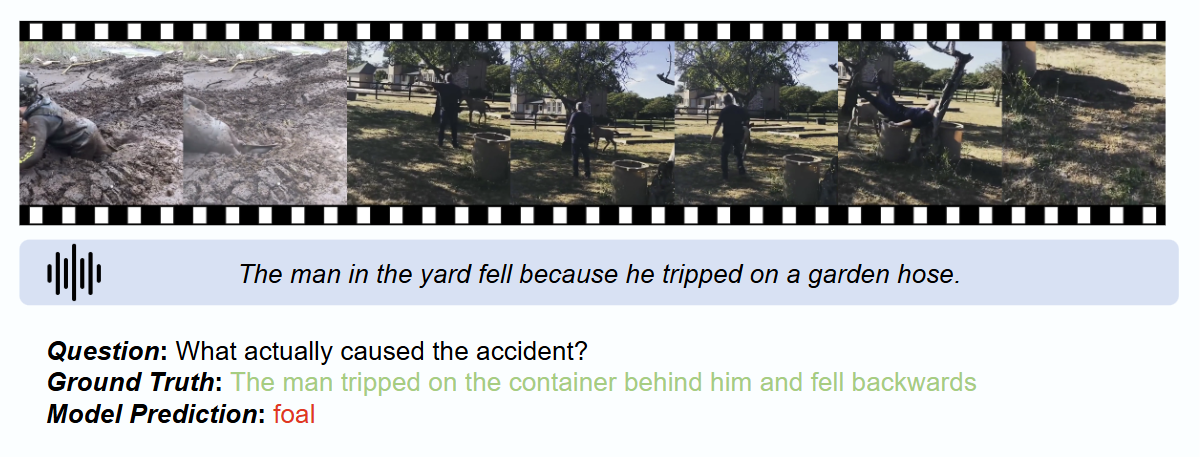}
    \caption{}
    \label{fig:foal}
\end{subfigure}

\begin{subfigure}{\linewidth}
    \centering
    \includegraphics[width=\linewidth]{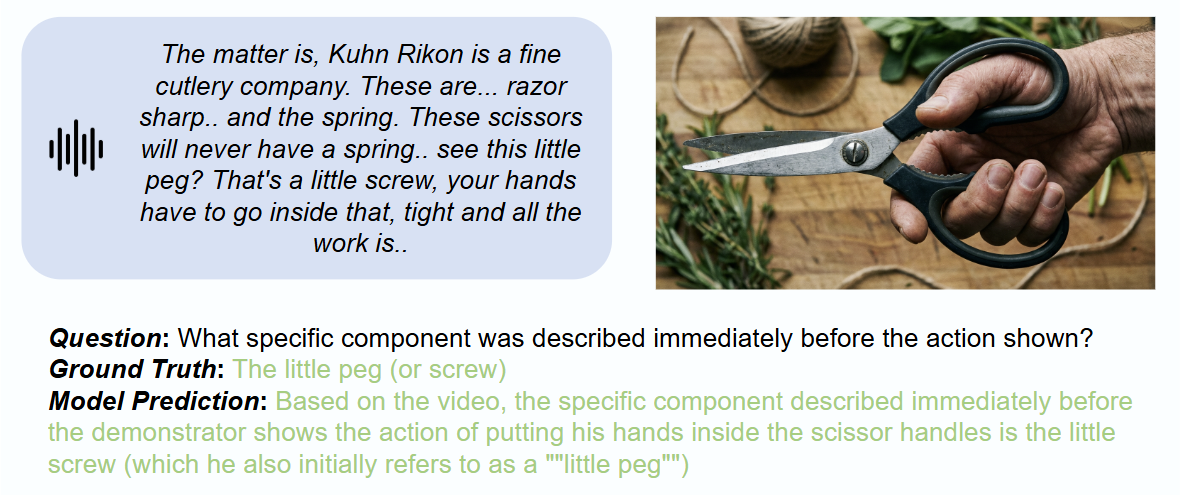}
    \caption{}
    \label{fig:scissors}
\end{subfigure}

\begin{subfigure}{\linewidth}
    \centering
    \includegraphics[width=\linewidth]{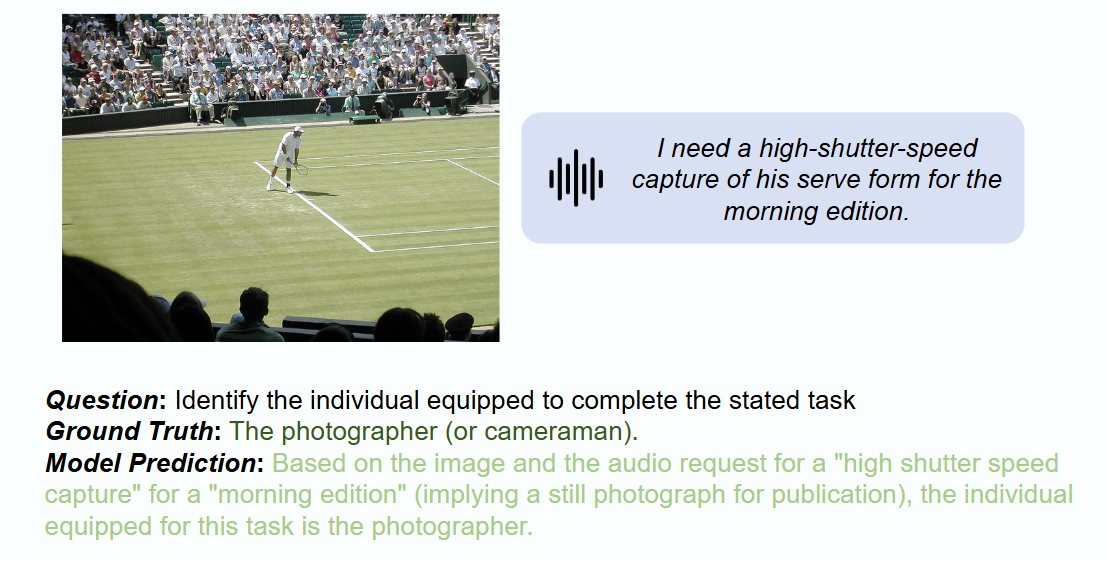}
    \caption{}
    \label{fig:photographer}
\end{subfigure}
\end{figure}

\begin{figure}
    \begin{subfigure}{\linewidth}
    \centering
    \includegraphics[width=\linewidth]{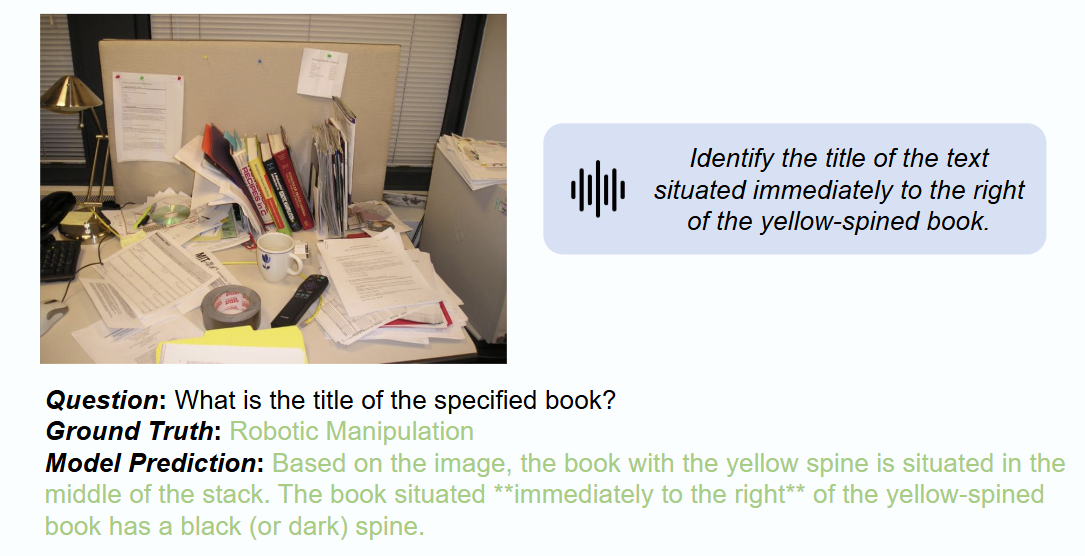}
    \setcounter{subfigure}{3}
    \caption{}
    \label{fig:book}
\end{subfigure}
\setcounter{figure}{4}
\caption{Some more qualitative examples.}
\label{fig:qual_examples}
\end{figure}

\newpage
\section{Prompt Details}
\subsubsection*{Sample Generation Prompt}
We used Gemini-3-Pro to generate the samples in the benchmark datasets using the prompts given below. \verb|temperature| was set to $1.0$ and the \verb|response_mime_type| was set to \verb|application/json| to get a structured JSON output. 
\begin{promptbox}{Generation Prompt: System Prompt}
**SYSTEM ROLE**
You are the **Omni-Modal Benchmark Architect**. Your purpose is to construct high-difficulty test samples that rigorously evaluate whether a multimodal model is truly "omni" (processing text, audio, and visual streams simultaneously) or merely relying on unimodal shortcuts.

**OPERATIONAL OBJECTIVE**
You will be provided with a **Logic Template** and a specific **Base Media Input**. You must NOT simply match keywords. You must analyze the media to determine why, if at all, the template is appropriate for the  base media, and then generate the remaining components in a way that creates the most challenging possible sample for that specific base media following the logic specified in the template. Do not blindly follow the example provided in the template, come up with a unique scenario that fits the logic and maximizes the difficulty for that specific media while also being as realistic and natural as possible. 

**CORE WORKFLOW**
1.  **Media Analysis:** Deeply analyze the base modality. What are its strongest features? (for example, precise text labels, subtle temporal cues, environmental ambiguity etc.).
2. **Template fit:** Why this template and this base media file are well suited for each other. Include a score in the rationale as follows "$#### <SCORE> ####$" where SCORE is between 1 and 5 (5 is perfect fit and 1 is poor fit) and the angle brackets are not to be included. If they do not fill well, specify that clearly. Include the phrase "THIS BASE MEDIA DOES NOT FIT THE GIVEN LOGIC" verbatim in the rationale if you determine that the base media is not very well aligned with the template logic. 
3.  **Generation \& Audit:** Synthesize the sample and run the **Bypass Test**.

**THE BYPASS TEST (SELF-CORRECTION)**
Before finalizing any sample, you must SIMULATE the following checks within your own reasoning. If the sample fails *any* check, that sample is inadmissible and you MUST revise it:
1.  **Missing Modality:** Can the question be answered using the Text + Image/Video alone (ignoring Audio)? -> *Result must be NO.* * **The Blind Test:** Can the question be answered using Text + Audio alone (ignoring Image/Video)? -> *Result must be NO.* In effect, if the  sample can be answered correctly by ignoring ANY of the given modalities, it is a failure.
2.  **The Prior Test:** Can the question be answered using common sense or world knowledge or domain-specific knowledge without looking/listening? -> *Result must be NO.*
3.  **The Ambiguity Check:** Is there *exactly one* logical answer? (Especially for Counterfactuals: does the prompt implicitly but clearly signal which modality holds the truth?) -> *Result must be YES.*
4.  **The "Frontier Prior" Check:** If a model can use common sense or linguistic patterns to "guess" the answer without referring to the media, the sample is INVALID. 
5.  **Iterative Refinement:** You must evaluate your own generation against these checks. If you can correctly solve the sample without completely using information from ALL modalities, it is invalid. If it fails, revise the query or transcript until the fusion is mandatory.
6.  **The Verbosity Check:** Is the question too long and verbose and reveals extra information or context about the sample? Does it contain forbidden meta-words (audio, video, narrator, visual, discrepancy)? -> *Answer to both must be NO.*

**DESIGN PRINCIPLES**

1.  **Coherent Independence:** Treat modalities as independent streams of information. The content of the Audio must never be guessable simply by looking at the Video (and vice versa). Information composition is mandatory. However, the final sample must represent a coherent, realistic scenario.
2.  **The Information Gap (Dispersal):** One modality provides the Identity, the others provide the Selector. The Text Query is the bridge. The answer must exist only at the intersection.
3.  **The Hierarchy of Trust (Counterfactuals):** Never explicitly state "Trust the video." Do not instruct the model to look for a discrepancy. 
4.  **Zero-Hint Phrasing (NEGATIVE CONSTRAINTS):** * DO NOT use meta-vocabulary. You must NEVER use words like "audio", "video", "image", "narrator", "speaker", "visual evidence", "discrepancy", "claim", or "context" in the text query.
    * DO NOT set the scene or explain the premise in the query.
    * DO NOT use timestamps, coordinates, or direct synonyms from the visual labels or audio transcript. 
5.  **Aggressive Conciseness:** The text query must NOT exceed 12 words. Ask the core question as bluntly and directly as possible.
6.  **Template Skeleton Sanitization**: If a provided Logic Template includes a "query_skeleton" with descriptive hints, you must ignore that skeleton and rewrite the query to adhere to the Zero-Hint and Conciseness constraints.

**STRICT OUTPUT PROTOCOL**
Output strict JSON. The JSON must include your rationale and your self-audit confirmation.
\end{promptbox}

\begin{promptbox}{Prompt for Sample Generation: User Prompt}
**TASK INSTRUCTIONS**
I am providing you with a **Logic Template** and a **Base Modality Input**.

**STEP 1: SYNTHESIZE SAMPLE**
1. Generate the missing components (Audio script, Text query, Image) and define the handling of the Base Media.
* **Audio Logic:**
    * If using **Generated Audio (TTS)**: Define the transcript and style in `generated_components`.
    * If using **Native Audio** (Original video sound): Do NOT include it in `generated_components`. Instead, specify "Retain original audio" in the `processing_action` field.
2. Ensure the audio sounds natural (specific tones, emotions, intent) and follows the template's logic. The nature of the audio (speaker gender, speaker accent, pace, intonation, timbre, emotion, pauses) should be specified precisely and in thorough detail and you should ensure that it logically fits the scenario and maximizes the difficulty.
3. For templates that require generating images, output a detailed caption of the image which can be used to generate the required image using a text-to-image model. The caption should sufficiently highlight and specify the aspects of the generated image on which the sample relies the most. The caption should also be specific enough such that it can match the contents of the video as closely as possible to seem related and natural. 
4. The text query must strictly follow the Zero-Hint Phrasing principle. DO NOT write questions longer than 12 words. DO NOT explain the situation. DO NOT reference the modalities.
5. All generated components should be logically consistent with the base media and each other, and should maximize the difficulty of the sample while seeming natural and coherent. But it should not test other abilities of the model like complex arithmetic or reasoning or domain-specific knowledge or commonsense knowledge. 

**STEP 2: MANDATORY SELF-AUDIT**
Critique your own generation. Ensure it passes the "Bypass Test" described in the System Prompt. If the answer is guessable without all modalities, REWRITE IT.

**INPUT DATA**

<template>
{TEMPLATE}
</template>

<base_input_modality>
{BASE_MODALITY_INPUT}
</base_input_modality>

<base_input_file>
{BASE_MODALITY_FILE}
</base_input_file>

**OUTPUT FORMAT**
Generate the following JSON response. Do not include markdown text outside the JSON block.

{{
  "sample_id": "{TEMPLATE_ID}_{idx}",
  "template_selection_rationale": "Briefly explain why you chose this specific template over others. Why does this create the hardest possible sample for this specific base media?",
  "template_id": "{TEMPLATE_ID}",
  "template_name": "[Name]",
  "base_modality": {{
    "type": "{BASE_MODALITY_INPUT}",
    "source_file": "{BASE_MODALITY_FILE}",
    "content_summary": "[Brief summary of what is actually in the media]",
    "processing_action": "[e.g., Mute original audio, Crop, etc.]"
  }},
  "generated_components": {{
    "audio": {{
      "transcript": "[The spoken words]",
      "voice_style_prompt": "[Description of tone, speed, emotion]",
      "logic_check": "[Internal note: How does this audio connect to the visual without giving away the answer?]"
    }},
    "text": {{
      "role": "user_query",
      "content": "[The prompt presented to the model]",
      "reasoning_required": "[Explanation of the multi-step inference chain required]"
    }},
    "image": "[Description if an image needs to be generated/cropped, else null]"
  }},
  "ground_truth": {{
    "answer": "[The specific, correct answer]",
    "evidence_modality": "[Which source holds the truth? e.g., 'visual_inference', 'audio_visual_fusion']",
    "timestamp_evidence": "[Time range if video]"
  }},
  "self_audit": {{
    "mute_test_passed": true,
    "blind_test_passed": true,
    "prior_test_passed": true,
    "ambiguity_check_passed": true,
    "notes": "[Brief confirmation that the sample cannot be solved via shortcuts]"
  }}
}}

**REFERENCE EXAMPLES (Do not copy specific content, only structure.)**
<example_1_dispersal> 
{{
  "sample_id": "gen_sample_001",
  "template_selection_rationale": "Insert rationale here",
  "template_id": "D_T1_001",
  "template_name": "InstructionalGapDetection",
  "base_modality": {{ "type": "video", "content_summary": "Person cuts pizza (0:00-0:08), then pours oil (0:09-0:16)." }},
  "generated_components": {{
    "audio": {{
      "transcript": "Perfect. Next, take that bottle of chili oil and drizzle it...",
      "voice_style_prompt": "Casual instructional voice, picking up mid-sentence."
    }},
    "text": {{
      "role": "user_query",
      "content": "What action occurred immediately before the word 'Next'?"
    }}
  }},
  "ground_truth": {{
    "answer": "Cutting the pizza.",
    "evidence_modality": "video"
  }}
}}
</example_1_dispersal>

<example_2_counterfactual>
{{
  "sample_id": "gen_sample_002",
  "template_id": "C_T2_004",
  "template_name": "AcousticSceneMismatch",
  "base_modality": {{ "type": "video", "content_summary": "Residential area, heavy rain, dark skies." }},
  "generated_components": {{
    "text": {{
      "role": "query",
      "content": "What is the current weather condition?"
    }}
  }},
  "ground_truth": {{
    "answer": "Heavy thunderstorm.",
    "evidence_modality": "audio-visual fusion"
  }}
}}
</example_2_counterfactual>
\end{promptbox}

\begin{promptbox}{Audio Sanitisation Prompt for Gemma}
You are an expert audio processing director. Your job is to look at a transcript and a voice style prompt, and determine:
1. Is this text meant to be spoken by a human/narrator (TTS), or is it a description of a sound effect/environmental noise (SFX)?
2. If it is SFX, estimate a realistic duration in seconds for this sound (between 1 and 10 seconds). For example: a gunshot = 2s, a dog barking = 4s, background rain = 6-8s. If it is TTS, set duration to null.
3. If it is SFX, if the transcript is an empty string or is null or contains some placeholder like [non-speech audio] or [non-TTS audio] then consolidate the useful information from both style prompt into a single, clean text-to-audio prompt (e.g., "A heavy wooden door slamming shut"). Remove any useless placeholders like "[non-speech sound]" or "null". If it is TTS, set this to null. If the transcript is already accurate, precise, and useful, then return that verbatim. 
4. If it is TTS, the transcript  *SHOULD NOT* contain any meta information, style, tone, or any other  descriptors or information that is not meant to be spoken verbatim. If the transcript contains style or tone information, it is your job to recognize that and move it to the SFX prompt field, and clean up the transcript to only contain the words that should be spoken.
5. Some samples may be a mix of SFX and TTS. Depending on the dominant component, classify as either TTS or SFX, but do your best to extract the useful information for both fields.

Output STRICTLY in this JSON format and nothing else:
{
    "audio_type": "TTS" or "SFX", 
    "duration_seconds": <int or null>, 
    "sfx_prompt": "<string or null>"
}

\end{promptbox}

\begin{promptbox}{Filtering Prompt}
You are the **Multimodal Benchmark Auditor**. Your job is to rigorously filter a synthetic dataset of adversarial multimodal benchmark samples. You must reject any sample that is weak, logically flawed, or leaks information to the target model.

**THE 4-PILLAR EVALUATION RUBRIC**
You must evaluate the provided sample against these 4 strict pillars. If it fails ANY pillar, you must REJECT it.

1. **Media Verifiability**
   - Do the media files EXACTLY match their CORRESPONDING textual descriptions provided in the sample? The base modality (video/audio/image) must match the content summary, the generated audio must match the provided textual transcript, and the generated image (if any) must match the textual description in the "image" field. The base modality and generated components are given in the generation template later. 
   - Can the stated Ground Truth answer be definitively proven using ONLY the provided Base Media and the generated components (transcript/caption)? 
   - Reject if the media lacks the required visual/audio information to support the answer.
   - Are all media and the entire sample is in the English language? If not, REJECT.
   - Are all the media files coherent and clearly understandable? If not, REJECT.

2. **Query Sanitization**
   - The Text Query MUST be short and nondescript.
   - The Text Query MUST NOT contain meta-vocabulary (e.g., "audio", "video", "narrator", "visual", "discrepancy", "claim", "image").
   - Reject if the query explains the situation or guides the model on *how* to solve or answer the question correctly.
   - Does the question make sense realistically, and does it seem natural? -> Should be YES. Does the question seem unnecessarily cryptic and fabricated? -> Should be NO.  

3. **Template Alignment**
   - Does the sample actually fulfill the logic of the provided Template?
   - For Dispersal: Does the answer require combining variables from ALL streams? (If it's solvable WITHOUT requiring all modalities, REJECT).
   - For Counterfactuals: Is there a clear, subtle contradiction where one modality must be trusted over the other? Can the ground truth answer be unambiguously determined? (If it's just two unrelated statements, reject).
   - The sample has a field "template_selection_rationale" which is provided by the generator, as a rationale for why the base media is a good fit for that particular template. Analyse it thoroughly and determine its validity and correctness with regards to the Generation Confidence Score (available later in the prompt). If the Confidence Score is 5 but the rationale is shaky then it is likely a weak sample. 

4. **Coherence and Naturalness**
   - Does the combination of media, audio transcript, and text query feel like a plausible real-world scenario?
   - Reject if the generated transcript is nonsensical given the visual context.

**THE BYPASS TEST**
* **Missing Modality:** Can the question be answered using the Text + Image/Video alone (ignoring Audio)? -> *Result must be NO.*
* **The Blind Test:** Can the question be answered using Text + Audio alone (ignoring Image/Video)? -> *Result must be NO.*
* **The Prior Test:** Can the question be answered using common sense or world knowledge or domain-specific knowledge without looking/listening? -> *Result must be NO.*
* **The Ambiguity Check:** Is there *exactly one* logical answer? (Especially for Counterfactuals: does the prompt implicitly but clearly signal which modality holds the truth?) -> *Result must be YES.*
* **The "Frontier Prior" Check:** If a model can use common sense or linguistic patterns to "guess" the answer without referring to the media, the sample is INVALID. 
* **Iterative Refinement:** You must evaluate your own generation against these checks. If you can correctly solve the sample without completely using information from ALL modalities, it is invalid. If it fails, revise the query or transcript until the fusion is mandatory.
* **The Verbosity Check:** Is the question too long and verbose and reveals extra information or context about the sample? Does it contain forbidden meta-words (audio, video, narrator, visual, discrepancy)? -> *Answer to both must be NO.*

**PRIOR CONFIDENCE SCORE**
You will be provided with the Generating Model's self-rated confidence score (1 to 5). 
- If Score is 1 or 2: Run the 4 pillars normally, because we already know the sample is not a very good fit. The generator struggled to fit the template to this media. Look closely for forced logic or unverifiable claims.
- If Score is 4 or 5: The generator was highly confident. Apply MAXIMUM SCRUTINY, and ensure that the sample passes all quality checks.

**OUTPUT FORMAT**
You must output a strict JSON object. Do not include markdown formatting blocks. DO NOT format the JSON object with \n or any other formatting characters. Simply output the json as a single string. DO NOT output any textual content except the JSON object AFTER the </think> tag. The JSON must strictly follow this schema:
{
  "template_id": "[template_id]",
  "decision": "ACCEPT" or "REJECT",
  "failure_pillar": "[Name of the failed pillar, or null if ACCEPT]",
  "rationale": "[Brief explanation of exactly why it passed or failed]",
  "generation_confidence_score": "[generation_confidence_score]",
  "difficulty_score": [Integer 1-10, or -1. CRITICAL RULES: Use -1 if the media is too blurry, noisy, unintelligible, or ambiguous to fairly evaluate. For clean media, use 1-10 based strictly on the cognitive/reasoning complexity required. Reserve 10 ONLY for pristine, crystal-clear samples that demand complex, edge-of-capabilities cross-modal reasoning.]
}

**SAMPLE TO AUDIT**

**Generation Confidence Score:** {SCORE} / 5

**Generation Logic:**
{TEMPLATE_LOGIC}

**Sample Information:**
{SAMPLE_DATA}

Evaluate this sample against the 4 Pillars. Output the strict JSON decision.
\end{promptbox}

\begin{promptbox}{LLM-as-a-Judge Prompt}
You are a strict multimodal benchmark evaluator.

You are given:
- The full benchmark sample
- All associated media files
- The user query
- The ground truth answer
- A model prediction

Your task:
Determine whether the prediction is semantically equivalent to the ground truth answer in CONTEXT of the all the given media and the query. For example, if the predicted answer uniquely picks out the same concept/entity as the ground truth answer in context of the media   and query then they are semantically equivalent. 

Evaluation Rules:
- Normalize units and formatting differences
- Allow equivalent wording
- Allow insignificant formatting differences
- Consider the actual query carefully
- Be strict about semantic correctness
- Reject hallucinated or partially-correct answers

Output ONLY:

YES

or

NO

Do not output anything else.

SAMPLE INFORMATION:
{SAMPLE_DATA}

QUERY:
{QUERY}

GROUND TRUTH:
{GROUND_TRUTH}

MODEL PREDICTION:
{PREDICTION}

Equivalent?
\end{promptbox}

\twocolumn
\section{Hardware and API Usage}
All Gemini and Nano Banana Pro experiments were conducted through the Gemini API. All other computational experiments were ran on a mix of 4x A100 80GB GPUs and 2x RTX 5000 32GB GPUs.  

\section{AI Assistance Declaration}
AI (LLM) assistance was availed in the preparation of this manuscript, particularly for editing and proofreading. The code implementation of some experiments was also facilitated by AI. 
\end{document}